\documentclass{article}

\PassOptionsToPackage{numbers, compress}{natbib}

\usepackage[main, final]{neurips_2026}

\usepackage{physics}
\usepackage[utf8]{inputenc} 
\usepackage[T1]{fontenc}    
\usepackage{hyperref}       
\usepackage{url}            
\usepackage{booktabs}       
\usepackage{amsfonts}       
\usepackage{nicefrac}       
\usepackage{microtype}      
\usepackage{xcolor}         
\usepackage{amsmath,bm}
\usepackage{amssymb}
\usepackage{algorithm}
\usepackage{algorithmic}
\usepackage{eqparbox}
\usepackage{graphicx}
\usepackage{pifont}

\title{Transporting Quantiles to Codebook: Scalable Vector Quantization without Codebook Collapse}

\author{%
  Yuwei Zeng\textsuperscript{1}\quad Zekun Shi\textsuperscript{1,2} \\
  \\
  \textsuperscript{1} National University of Singapore \textsuperscript{2} SEA AI Lab
}

\begin{document}

\maketitle

\newif\ifchecklist
\checklistfalse

\begin{abstract}

Effective discrete representation learning with VQ-VAE serves as a fundamental component of modern autoregressive generative models. However, it is widely known to suffer from codebook collapse and inefficient code utilization. To address this issue, we propose FlowVQ, a simple yet effective alternative to standard vector quantization by formulating neural quantization as a joint modeling and clustering process. Specifically, we learn the latent space using a lightweight flow matching model alongside training, corresponding to a probability flow ODE that transports an often fixed well-behaved source distribution while preserving probability mass. Based on this, we define the codebook $\mathcal{C}_1$ by transporting even-quantile partitioned prior codes $\mathcal{C}_0$ from the source space through the learned flow, and obtaining the Voronoi centroids under the induced distribution, leading to a structured discretization. We evaluated on image tokenization and generation, as well as on two tasks requiring high-precision tokenization, demonstrating scalability and improved expressivity of the codebook with full utilization. Code is available at: \href{https://github.com/friolero/flowvq}{\textcolor{blue}{GitHub repository}}.
    
\end{abstract}
\section{Introduction}

Vector Quantized Variational AutoEncoder (VQ-VAE), proposed by~\citet{van2017neural},~\cite{razavi2019generating} demonstrates effective discrete latent representation learning by incorporating Vector Quantization (VQ) into reconstruction. It learns a finite set of codebook vectors and tokenizes continuous data into discrete sequences by mapping to the nearest vectors, enabling large autoregressive models for continuous data generation while leveraging the design, training, and scaling principles behind large language models. By balancing fidelity, expressiveness, and computational efficiency, VQ-VAE remains an essential building block of modern generative models that are adopted across diverse domains, including state-of-the-art models for image generation and video-based world modeling.

With codebook collapse and unscalable model capacity remaining unresolved, existing works have explored numerous ways to mitigate such issues. Early improvements~\cite{dhariwal2020jukebox, agustsson2017soft, zheng2023online} involved various training techniques such as random restart, random replacement, and k-means initialization. ~\citet{williams2020hierarchical} and~\citet{takida2022sq-vae, takida2024hq} proposed stochastic quantization,~\citet{zhu2024scaling} and~\citet{chang2026scalable} proposed adding an additional projector layer or bridging network after a frozen codebook. With an altered latent space, model performance can be compromised, leading to degenerate reconstruction and representation learning. Recently, Lookup-Free Quantization (LFQ)~\cite{mentzerfinite, yu2024language, webermaskbit} methods proposed an alternative to VQ based on direct scalar quantization with fixed hypercube geometry, rounding values in each dimension to the nearest point, and showed scalable codebook capacity with improved image generation using large vocabulary sizes beyond $2^{14}$. 

To circumvent codebook collapse, we propose that an ideal codebook should jointly model and cluster the latent space into even partitions alongside reconstruction training. The codebook vectors correspond to the Voronoi centers of these partitions, naturally allocating denser vectors to high-density regions and sparser vectors to low-density regions, regardless of the manifold geometry. This ensures full and balanced utilization, addressing expressiveness and scalability. Based on this principle, we propose a flow-based vector quantization approach without interfering with the representation learning objective. 

\begin{figure}[t]
    \centering
    \includegraphics[width=\textwidth]{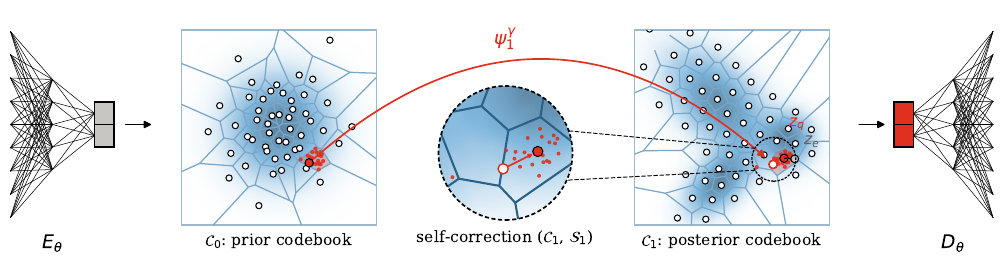}
    \caption{FlowVQ on a three-mode 2D latent distribution (K = 50): the learned flow $\psi^\gamma_1$ transports the approximately equal-mass quantile prior codebook $\mathcal{C}_0 \in \mathcal{N}(\mathbf{0}, I)$, initialized using Sobol samples here and kept fixed thereafter, to the latent space. Self-correction refines each transported code using the pushed quantile samples $\mathcal{S}_1$, yielding the density-adaptive posterior codebook $\mathcal{C}_1$.}
    \label{main_diagram}
\end{figure}

A flow matching model~\cite{lipman2022flow} learns a continuous transport mapping samples from a prior distribution, such as a Gaussian, to an arbitrary data distribution. This transport corresponds to a Probability Flow ODE (PF-ODE)~\cite{maoutsa2020interacting, song2020score}, which preserves probability mass and transports prior quantiles to data quantiles. Based on this, we define the prior codebook uniformly in the well-behaved source space, and co-train the flow model with VQ-VAE. The codebook vectors are then obtained as the pushforward of the prior codebook and move toward the Voronoi centers under the estimated density. Experiments show that a lightweight flow model introduces minor overhead relative to VQ-VAE while maintaining full code utilization and improving performance.

In summary, our contributions are as follows:
\begin{enumerate}
    \item We propose FlowVQ, a novel VQ technique that converts codebook learning into a quantile transport problem, where code utilization emerges as a geometric property of the latent space, enabling balanced usage and avoiding codebook collapse.
    \item We show that FlowVQ consistently achieves full codebook utilization across a range of tasks, improving reconstruction and generation quality within a unified tokenizer framework.
\end{enumerate}
\section{Related Works}
\label{sec: related_works}

\paragraph{VQ-VAE \& Generation in the Latent Space} Proposed by~\cite{van2017neural}, the VQ-VAE combines vector quantization with a reconstruction objective and achieves effective neural tokenization. It enables autoregressive generation on continuous data, such as high-fidelity image generation, by predicting the sequence of tokens in the latent space using a PixelCNN~\cite{van2017neural, van2016conditional} and, more recently transformer models. VQ-GAN~\cite{esser2021taming} further couples the training with a discriminative loss, addressing the issue of blurry reconstruction from traditional VAE greatly improving generation fidelity. It has since been adopted as a standard framework. Recently, advancements on residual~\cite{lee2022autoregressive}, multi-head~\cite{zheng2022movq}, and multi-scale~\cite{tian2024visual} tokenization were proposed and demonstrate solid improvements~\cite{lee2022autoregressive, tian2024visual, zhang2024preventing}. As a general tokenization technique, VQ-VAE based tokenizers are widely adopted in diverse generative tasks today, especially in tasks where direct generation can be prohibitively expensive such as image~\cite{yu2024language, tian2024visual, rombach2022high}, audio~\cite{dhariwal2020jukebox, dieleman2018challenge}, and video~\cite{yan2021videogpt, bruce2024genie}, or in fields where structural representations are often unavailable, such as AI for protein and material generation~\cite{pourmirzaei2025gcp, lu2025uni, qiu2024vqcrystal}.

\paragraph{Mitigating Codebook Collapse} To address the codebook collapse, which leads to unscalable codebook capacity, an active line of work has been proposed to mitigate it. Early work addressed it with various tricks such as random restart or random replacement~\cite{dhariwal2020jukebox, zheng2023online}. Another line of work explored differentiable relaxation~\cite{agustsson2017soft, baevskivq, jiang2026differentiable}, most notably using the Gumbel–Softmax~\cite{jang2017categorical} . However, these approaches generally underperform and produces lower-quality reconstruction~\cite{williams2020hierarchical}. Takida et al.~\cite{takida2022sq-vae, takida2024hq} proposed improving the mode-dropping behavior by switching to stochastic quantization but suffer from similar unrealistic reconstructions. \citet{zhu2024scaling} and~\citet{chang2026scalable} addressed the under-utilization issue by adding intermediate projection layer(s) between a frozen codebook and the decoder. In a parallel track,~\citet{mentzerfinite},~\citet{yu2024language}, and~\citet{webermaskbit} proposed LFQ, which rounds each dimension to the nearest point in a hypercube, and scales the codebook capacity with its size using direct scalar quantization. These methods demonstrate competitive or improved image generation performance, especially with large vocabulary sizes beyond $2^{14}$ to $2^{18}$. 

The majority of the existing works mitigate this issue without fundamentally solving it. To our knowledge, the closest work is OptVQ~\cite{zhang2024preventing},  which introduces global information via optimal transport between batch latents and the codebook. However, its cost matrix requires per-batch normalization for numerical stability, which distorts the transport cost inconsistently across batches. Moreover, the resulting doubly stochastic assignment is converted to a hard assignment, which does not guarantee that every code is selected, especially under stronger entropy regularization. Full utilization thus depends on the entropy weight, which trades off quantization accuracy against utilization and may be sub-optimal for a given task as observed in our experiments.
\section{Method}

\subsection{Preliminary}

We define a codebook $\mathcal{C} = \{\textbf{C}_i\}_{i=1}^K$ where $\textbf{C}_i \in \mathbb{R}^{d}$, K is the number of code vectors and d is the latent vector dimension. Given an input $x$, the original latent vector is $z_e = E_\theta(x)$, and the quantized vector $z_q$ is defined as: 
\begin{equation}
    z_q = \textbf{C}_i \text{\quad where \quad } i = \arg\min_{i} ||\textbf{C}_i - z_e||_2
\end{equation}

With sg() indicating the stop gradient operator, the original VQ-VAE trains with the following objective function. $\beta$ is a training dynamics dependent coefficient that controls the expansion rate of $z_e$ driven by the reconstruction objective, while ensuring the effectiveness of the STE operator in the reconstruction loss:
\begin{equation}
    \mathcal{L}_\text{VQ-VAE}(\theta, \mathcal{C}) \quad=\quad\underbrace{||D_\theta(\text{STE}(z_q)) - x||_2^2}_\text{reconstruction loss} \quad + \quad\underbrace{||\text{sg}(z_e) - z_q||_2^2}_\text{codebook loss}\quad +\quad \beta \underbrace{||z_e -\text{sg}(z_q)||_2^2}_\text{commitment loss}
    \label{eq: vqvae_loss}
\end{equation}
\[\text{where\quad} \text{STE}(z_q) = z_e + \text{sg}(z_q - z_e)\]

\subsection{Flow-based Codebook}
\label{sec: codebook_update}
In this work, we propose to define the codebook in the prior space, which we call the prior codebook $\mathcal{C}_0$. The prior space can be arbitrary, as long as it supports uniform sampling. To achieve uniform sampling, we partition the prior space into K equal probability mass quantiles using its cumulative distribution function (CDF) $\mathcal{F}$. $\{\mathbf{C}_0^i\}_{i=1}^K$ is then obtained by applying the inverse CDF $\mathcal{F}^{-1}$ to the centers of these quantiles:
\[\textbf{C}_0^i = \mathcal{F}^{-1}\left(\dfrac{i-0.5}{K}\right), \quad i=1,..., K\]
In higher dimensional latent spaces, as d increases, exact uniform quantile partitioning becomes intractable. We use Sobol sequences, a low-discrepancy quasi-random sequence designed to evenly fill multi-dimensional spaces to generate approximate uniformly distributed quantile centers. Specifically, we first generate K points in the CDF as d-dimensional unit hypercube $[0, 1]^d$:
\begin{equation}
    \{\textbf{u}_i\}_{i=1}^K = \text{Sobol}(K, d)\quad \text{where}\quad \textbf{u}_i \in [0, 1]^d
\end{equation}
Then obtain the prior codebook for each quantile similarly
\begin{equation}
    \textbf{C}_0^i = \mathcal{F}^{-1}(\textbf{u}_i) 
\end{equation}

We define the learned quantizer flow velocity field as $u_t^\gamma$, and the associated flow map as $\psi _t^\gamma$. Let the data distribution be $x\sim p_{\text{data}}$ . During VQ-VAE training, it is trained to match the current latent distribution over $z_e$ with the conditional flow matching objective:
\begin{equation}
    \mathcal{L}_\text{FM}(\gamma) = \mathbb{E}_{t \sim \mathcal{U}[0, 1],\  z_0 \sim \mathcal{N}(0, I_d), \  x \sim p_{\text{data}}}||u_t^{\gamma}(z) - u_t(z\mid z_1, z_0)||_2^2
\end{equation} 
\[ \text{where}\quad\dfrac{d}{dt}\psi_t^\gamma(z) = u_t^\gamma (\psi_t^\gamma(z)),\quad z_1=E_\theta(x), \quad \psi_t(z\mid z_1, z_0) = (1-t)z_0 + tz_1\]

With latent distribution modeling and clustering handled by a separate flow model, the explicit codebook loss term in Eq  (\ref{eq: vqvae_loss}) is no longer required. The flow matching objective is optimized in a decoupled manner and should introduce minimal interference to the reconstruction and quantization objectives. The overall training objective is then updated to:
\begin{equation}
    \begin{aligned}
        \mathcal{L}_\text{FlowVQ}(\theta,\gamma) & =  \mathcal{L}_\text{VQ-VAE}(\theta)\ +\ \mathcal{L}_\text{FM}(\gamma ) \\
        &= ||D_\theta(\text{STE}(z_q)) - x||^2\ +\ \beta ||z_e - \text{sg}(z_q)||^2\ +\  \mathbb{E}_{t, z_0, z_1}||u_t^{\gamma}(z) - u_t(z\mid z_1, z_0)||^2 
    \end{aligned}
    \label{eq: loss_flow_vq}
\end{equation}

The final codebook $\mathcal{C}_1$ used in vector quantization is periodically updated. Each code vector $\textbf{C}_1^i$ is obtained as the pushforward of the prior codebook $\textbf{C}_0^i$ through the trained flow $\psi_t^\gamma$:
\begin{equation}
    \textbf{C}_1^i = \psi_1^\gamma(\textbf{C}_0^i), \quad i = 1, ..., K
    \label{eq: c1_pushforward}
\end{equation}
The codebook structure closely follows the geometry of the latent manifold and adapts to its density, naturally providing sharper resolution in denser regions and coarser in sparser regions. This is in contrast to the original VQ, which is sensitive to both initialization and training dynamics due to its reliance on both codebook and commitment losses between only matched pairs, which are a root cause of codebook collapse. Similarly, fixed-geometry approaches such as LFQ do not adapt to the latent distribution, including empty regions beyond the latent boundaries and holes within it, which limits their expressivity.

\subsection{Self-Correction toward Voronoi Centroids}
\label{sec: codebook_optim}

Up to this point, a set of approximately uniform points in the latent space is obtained through Eq (\ref{eq: c1_pushforward}) as the codebook $\mathcal{C}_1$, which only fulfills the first two key properties of faithful latent modeling and approximately even partitioning. To minimize the approximation error arising from the quasi-random sequence and the non-linear flow mapping, we introduce a self-correction mechanism that updates $\mathcal{C}_1$ toward the Voronoi centroids as the Euclidean centers of the transported quantiles, aiming to achieve the final desired property of the codebook. \emph{This process can be viewed as transporting each quantile cell through a set of its uniformly spread samples.} 

The trained flow serves as a proxy for the latent data distribution. Denoting N as a sampling ratio, we uniformly draw NK points $\{\textbf{S}_0^i\}_{i=1}^{NK}$ from the prior and push them through the flow to the latent space as dense samples $\mathcal{S}_1$ similarly, simulating the incoming latent data for vector quantization as $\{z^{'}_e\}$. 
\begin{equation}
\begin{aligned}
    \mathbf{S}_0^i & = \mathcal{F}^{-1}(\textbf{v}_i), \quad \text{where}\quad \{\textbf{v}_i\}_{i=1}^{NK} = \text{Sobol}(N K, d), \ \textbf{v}_i \in [0, 1]^d \\
    \textbf{S}_1^i & \leftarrow \psi_1^\gamma(\textbf{S}_0^i), \quad i = 1, ..., NK
\end{aligned}
\end{equation}

We then cluster $\{\textbf{S}_1^i\}_{i=1}^{NK}$ with respect to the current codebook $\mathcal{C}_1$ and update each $\textbf{C}_1^i$ toward the corresponding cluster centroid, implementing a self-correction mechanism that improves codebook alignment with the latent manifold. This process is analogous to k-means clustering initialized with well-chosen centers.
\begin{equation}
    \mathcal{C}_1 = \arg\min_{\mathcal{C}_1}\sum_{j=1}^K\sum_{\textbf{S}_1^i\in Q_{j}}||\textbf{S}_1^i - \textbf{C}_1^j||_2^2
    \label{eq:online_correction}
\end{equation}

Where $\{Q_j\}_{j=1}^K$ denotes the K quantile clusters in the latent space, and $\textbf{C}_1^j$ corresponds to the current codebook vectors of the quantiles. This process is performed iteratively a few times and deterministic with minimal overhead.

Optionally, we can further correct the codebook $\mathcal{C}_1$ online towards the cluster centroids of real latent data $\{z_e\}$ during co-training. This could reduce the remaining approximation error due to finite samples and model error. Specifically, this online correction reinstates the codebook loss of $\mathcal{L}_\text{VQ-VAE}$ in Eq (\ref{eq: vqvae_loss}) in Eq (\ref{eq: loss_flow_vq}), which can pose a risk of re-introducing codebook collapse between two codebook updates. However, due to the quality of the flow partition and frequent codebook update, we did not observe significant collapse in our experiments and found it can further improve the model slightly over the purely self-corrected version.    

\section{Experiments}

In this section, We evaluate FlowVQ on image reconstruction and generation in terms of generation quality with common metrics such as FID, IS, and codebook quality, measured by its entropy and utilization rate. We first evaluate it on CIFAR-10~\cite{krizhevsky2009learning} in Sec~\ref{eval_cifar10}, followed by ImageNet-1K~\cite{deng2009imagenet} at 256$\times$256 resolution in Sec~\ref{eval_imagenet}. 

To further examine the expressivity induced by FlowVQ and its general applicability, we evaluate FlowVQ on two high-precision tasks: crystalline material tokenization in Sec~\ref{eval_mattok}, and action tokenization for robotics in Sec~\ref{eval_acttok}

\subsection{Image Tokenization}

\subsubsection{CIFAR-10}
\label{eval_cifar10}

As various design choices affect the final model performance, to understand the sole effect of VQ techniques, we first conduct a set of controlled experiments on CIFAR-10. All models share the same encoder and decoder architecture and training scheme (47M parameters), differing only in the quantizer, i.e., quantization and codebook update. The image is compressed to the latent space by a fixed factor $f = 4$, i.e. from $H \times W $ to an $H/f \times W/f$ patch token grid. We evaluate reconstruction under different codebook sizes and dimensions to examine the codebook scalability. 

For baselines, we compare against two most common quantization methods: VQ-GAN~\cite{esser2021taming, webermaskbit}, and hypercubic LFQ method~\cite{yu2024language, webermaskbit}. We also compare with OptVQ~\cite{zhang2024preventing}. For a fair comparison with LFQ method , which ties the codebook size $|\mathcal{C}|$ to the latent vector dimension $d: |\mathcal{C}| = 2^{d}$, we adjust d accordingly for $|\mathcal{C}| = 1024$ and $|\mathcal{C}| = 16384$. We also explored the effect of auxiliary entropy loss, a technique widely adopted to improve codebook utilization in existing works~\cite{mentzerfinite, yu2024language, webermaskbit, chang2022maskgit}, which optimizes the mutual information between the latent inputs and codebook, as illustrated in~\cite{krause2010discriminative, jansen2020coincidence}: 
\begin{equation}
    \mathcal{L}_{ent} = \mathbb{E}[H(q(z))] - H[\mathbb{E}(q(z))]
    \label{eq: ent_loss}
\end{equation}

\begin{table}[H]
\centering
\resizebox{\textwidth}{!}{    
    \begin{tabular}{c|cc|cc|ccc|c|cc}
        \toprule
         & d & $|\mathcal{C}|$ & rFID$\downarrow$ & IS $\uparrow$& SSIM$\uparrow$ & PSNR$\uparrow$ & LPIPS$\downarrow$ & MAE$\downarrow$ & Entropy$\uparrow$ & Util $\%\uparrow$ \\
        \midrule
        \textit{VQ-GAN+} & 10 & 1024 & 2.40  & 10.77 & 0.78 & 23.62 & 0.130 & 0.050 & 8.81 & 45.6\\
          & 10 & 10240 & 2.65 & 10.70 & 0.75 & 23.27 & 0.143 & 0.052 & 8.07 & 2.7\\
        \small{w/o $\mathcal{L}_{ent}$} & 14 & 16384 & 1.93 & 10.88  & 0.81 & 24.79 & 0.115 & 0.044 & 8.71 & 2.6 \\
        & 14 & 16384 & 1.65 & 10.87  & 0.81 & 24.82 & 0.114 & 0.044 & 8.57 & 2.8\\
        \midrule
        LFQ & 10 & 1024 & 3.00  & 10.65 &  0.74 & 23.01 & 0.149 & 0.058 & 8.46 & 50.0 \\
        \small{w/o $\mathcal{L}_{ent}$} & 14 & 16384 & 2.95 & 10.83 & 0.73 & 22.57 & 0.161 & 0.056 & 5.91 & 0.8 \\
         & 14 & 16384 & 2.03 & 10.83  & 0.83 & 24.96 & 0.104 & 0.043 &  13.83 & 100.0  \\
        \midrule
        OptVQ & 10 & 1024 & 3.42  & 10.79 & 0.76 & 23.50 & 0.138 & 0.050 & 9.99 & 100.0 \\
         & 10 & 10240 & 1.95  & 10.88 & 0.85 & 25.31 & 0.091 & 0.041 & 13.27 & 100.0 \\
         & 14 & 16384 & 1.72 & \textbf{10.93} &  0.85 & 25.78 & 0.087 & 0.039 & 13.94 & 100.0 \\
        \midrule
        FlowVQ & 10 & 1024 & 2.11 & 10.71 & 0.80 & 24.28 & 0.118 & 0.046 & 9.93 & 100.0 \\
          & 10 & 10240 & 1.87 &  10.68  & 0.83 & 25.39 & 0.100 & 0.040 & 13.28 & 100.0 \\
         & 14 & 16384 & 1.40 & 10.72  & 0.86 & 26.50 & 0.086 & 0.036 & \textbf{13.97} & 100.0 \\
         \small{w/ $\mathcal{L}_{ent}$}& 14 & 16384 & \textbf{1.39} & 10.89 & \textbf{0.87} & \textbf{26.72} & \textbf{0.081} & \textbf{0.035} & \textbf{13.97} & 100.0 \\
        \bottomrule
    \end{tabular}
}
\caption{Evaluation of image quality and codebook utilization on CIFAR-10. By default, all baselines optimize the auxiliary entropy loss while FlowVQ does not if not specified. As the entropy maximization is tightly coupled in the OptVQ optimization process, we did not explore removing it.}
\label{tab:cifar10}
\end{table}

For FlowVQ, the velocity field is modeled using a lightweight 4-layer MLP with hidden dimension $h = 256$ (0.2M extra parameters). We observe this setting to be effective in producing an aligned latent space and fix it across all tasks. We set the source distribution to a normal distribution $\mathcal{N}(\mathbf{0}, \mathbf{I}_d)$. The dense prior samples are generated with a ratio N = 50. We start the initial codebook construction at the $1000^{th}$ step, then update it every 1000 steps using batch size 128. 

\paragraph{Result Analysis} The results are summarized in Table~\ref{tab:cifar10}. We can spot the codebook collapse of the original VQ immediately when $|\mathcal{C}|$ increases from 1024 to 10240 with a fixed latent dimension. The number of active codes decreases from $1024 \times 45.6\% = 467$ to $10240 \times 2.73\% = 280$, leading to degenerate overall performance. The entropy regularization drastically improves the utilization rate of LFQ, but does not guarantee full utilization when balancing multiple objectives of minimizing quantization error and maximally filling the hypercube simultaneously. The same conflict applies to OptVQ, which was observed to learn fast in early steps, but plateaus much earlier than methods that are solely driven by the quantization objective, such as VQGAN and FlowVQ here. Nevertheless, we note that VQ-GAN serves as a strong baseline due to its flexibility, which leads to high-quality alignment between the codebook and an evolving latent space. 

We observe that FlowVQ achieves significant improvements in both image quality, with rFID = 1.40, reconstruction error MAE of 0.036, and codebook quality with full utilization across all settings. The performance consistently improves with increasing $|\mathcal{C}|$. Explicit optimizing the mutual information further improves the model slightly, though we note the performance is very close between the two. Both variants achieve the highest entropy (13.97, near the maximum $log_2{16384}=14$), indicating the most balanced usage across codes, which validates the effectiveness of the proposed method.

\paragraph{Codebook Visualization}

\begin{figure}[H]
    \centering
    \includegraphics[width=\textwidth]{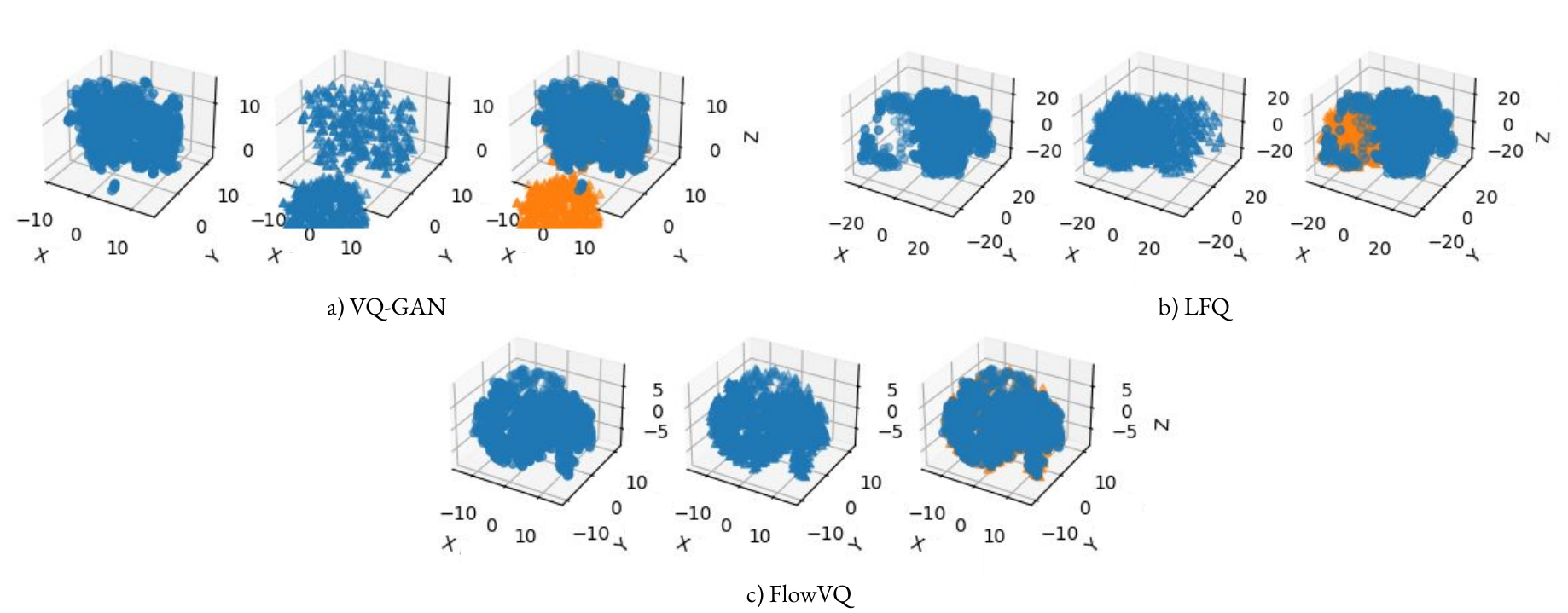}
    \caption{Visualization of \textbf{\textit{(left)}} the latent vectors on the evaluation set {\Large $\bullet$}; \textbf{\textit{(middle)}} the codebook $\blacktriangle$; \textbf{\textit{(right)}} the joint plot of \textcolor{blue}{latent vectors {\Large$ \bullet$}} and \textcolor{orange}{codebook vectors $\blacktriangle$} in each plot. For each model, the latent vectors and code vectors are randomly down-sampled to the same number, then jointly reduced to 3D using t-SNE for visualization.}
    \label{codebook_vis}
\end{figure}

We next provide visualizations of the latent space on the evaluation set and codebook alignment between FlowVQ and baselines without global information in Fig~\ref{codebook_vis}. We can see the codebook collapse pattern in VQ-GAN, where a large number of codes at the bottom left remain unused despite the active codes aligning well with the latent space, while LFQ imposes a hypercube geometry, so holes in latent space inevitably lead to unused codes. FlowVQ, on the contrary, models the geometry flexibly similar to VQ-GAN and imposes no additional regularization on the latent shape, resulting in high code utilization that allows it to scale with the codebook size. 

\paragraph{Time Analysis}\label{time_analysis} One potential concern for the proposed method is that the pushforward operation can be time-consuming. We measure the average training time per 1000 steps (one codebook update cycle) for all models in robotics VLA model Table~\ref{tab:time_analaysis} in Appendix~\ref{app_train_time}. FlowVQ takes on average 618s compared to 535s for VQ-GAN, increased by 15.6\% with the difference mostly due to the codebook update, which takes 82.8s on average. This time is expected to further decrease with more GPU memory. Meanwhile, as the prior vectors are pre-defined in the fixed source space, the update is fully deterministic and easily supports distributed execution with minimal communication overhead. We implement the distributed update. It takes 3s per codebook update on 32 NVIDIA A100 GPUs for $|\mathcal{C}|=16384$ for large scale training such as ImageNet, thus comparable speed to the original VQ-GAN.

\subsubsection{ImageNet-1K}
\label{eval_imagenet}

\begin{figure}[H]
    \centering
    \includegraphics[width=\textwidth]{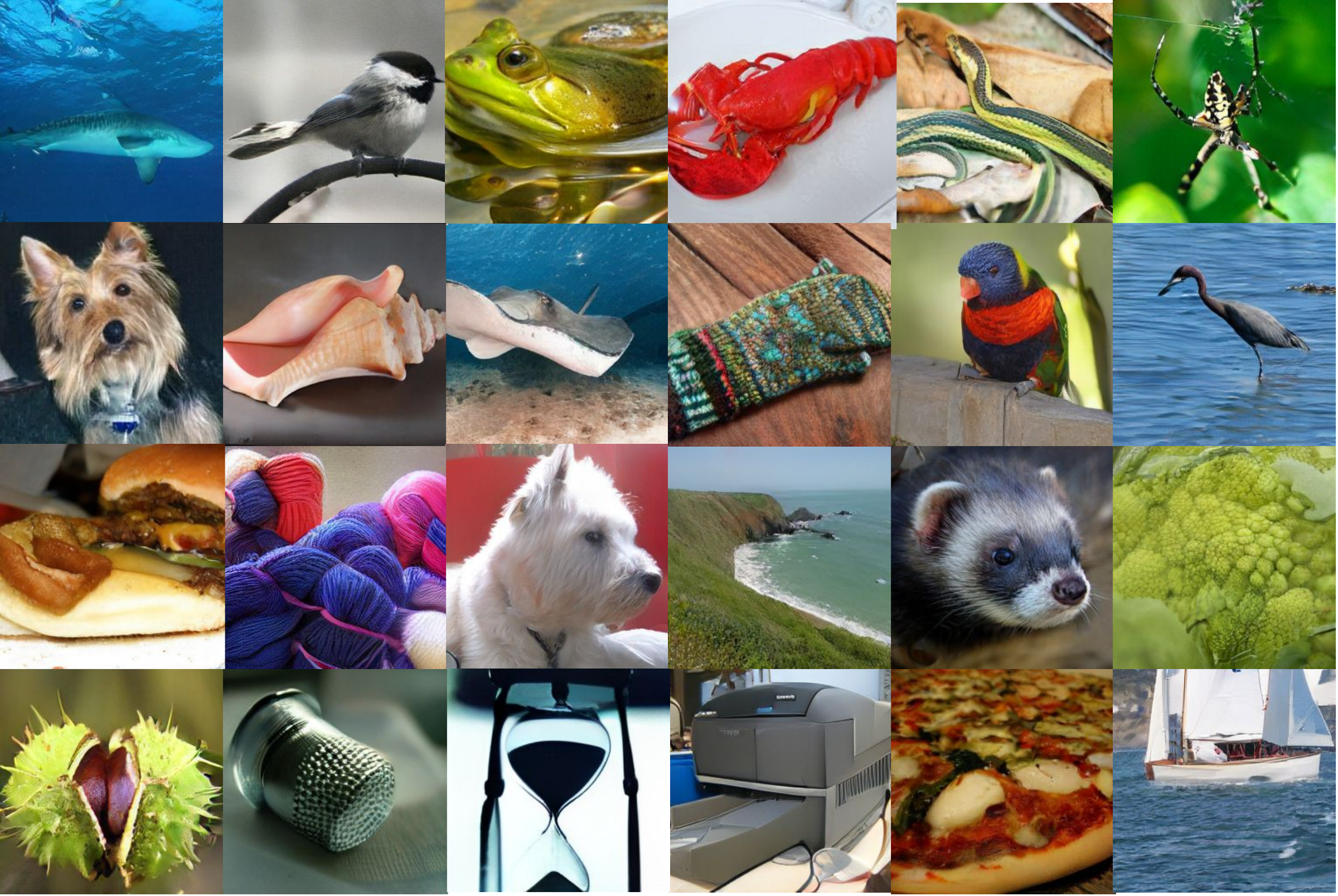}
    \caption{class-conditioned generation on ImageNet-1K $256\times256$ trained model. Each image corresponds to a different class. More samples can be found in Appendix~\ref{app_gen_samples}.}
    \label{fig: imagenet_gen}
\end{figure}

\begin{table}[H]
\centering
\resizebox{\textwidth}{!}{    
    \begin{tabular}{c|ccc|c|ccc}
        \toprule
         \textbf{Model}  & \textbf{Tokenizer} & \textbf{$|\mathcal{C}|$} & \textbf{\#tokens}  & \textbf{\#params} & \textbf{rFID}$\downarrow$  & \textbf{gFID} $\downarrow$ & \textbf{IS} $\uparrow$  \\
        \midrule
        \textit{ImageNet-1K val. data}~\cite{tian2024visual} &  & & & & & 1.78 & 236.9\\
        \midrule
        LDM-4-G~\cite{rombach2022high} & Continuous$^{*}$ & 8192 & $64^2$ & 400M & - & 3.60 & 247.7  \\
        LDM-8-G~\cite{rombach2022high} & Continuous$^{*}$ & 16384 & $32^2$ & 506M & -  & 7.76 & 209.5  \\
        MAR-L-407M~\cite{li2024autoregressive}  & Continuous & - & $16^2$ & 407M  & 1.22  & 1.97 & 291.2 \\
        \midrule
        FSQ~\cite{mentzerfinite}     & LFQ & 1000  & $16^2$ & 155M  & -    & 4.53 & -  \\
        MAGVIT-v2~\cite{yu2024language} & LFQ & 262144   & $16^2$ & 307M & -  & \textbf{1.78} & 319.4 \\
        Open-MAGVIT2-AR-B~\cite{luo2024open} & LFQ & 262144  & $16^2$ & 343M & 1.67 & 3.08 & 258.3 \\
        BSQ~\cite{zhaoimage}  & LFQ & 262144 & $16^2$ & 174M & \textbf{1.14} & 5.44 & 139.6 \\ 
        \midrule
        taming-VQGAN & VQ & 1024 & $16^2$ & 310M & 7.94 & 5.88 & 304.8 \\
        taming-VQGAN & VQ & 16384 & $16^2$ & 310M & 4.98 & -  & - \\
        MaskGIT~\cite{chang2022maskgit}  & VQ & 1024 & $16^2$ & 227M & 2.28  & 6.18 & 182.1 \\
        SQ-VAE-2~\cite{takida2022sq-vae} & VQ & 3072 & $16^2$ &- & 4.51 & - & - \\
        LlamaGen~\cite{sun2024autoregressive} & VQ & 16384 & $16^2$ & 343M & 2.19 & 4.21 & 200.1\\
        VAR-d16~\cite{tian2024visual} & VQ & 4096 & $16^2$  & 310M & -  & 3.30 & 274.4 \\
        VAR-d20~\cite{tian2024visual} & VQ & 4096 & $16^2$ & 600M  & -    & 2.57 & 302.6 \\
        \midrule
        FlowVQ & VQ & 16384 & $16^2$ & 307M & 1.48  & 1.89 & \textbf{351.4} \\
        \bottomrule
    \end{tabular}
}
\caption{Image reconstruction and generation on ImageNet-1K $256\times256$ with classifier free guidance.  We compare FlowVQ with the existing works with similar model capacity and their guided or rejection sampling scores. {\footnotesize* LDM reconstructs on continuous space, but generation discretizes input to the nearest token.}}
\label{tab:imagenet_1k}
\vspace{-0.5pt}
\end{table}

We next extend the evaluation on ImageNet-1K at a resolution of $256\times 256$ for both reconstruction and generation. For stage one, the model shares a similar design to Sec~\ref{eval_cifar10} but is constructed deeper using channel multiplier as [1,1,2,2,4], leading to a compression ratio $f = 16$ and a $16 \times 16$ token grid. We train the model for 1.35M steps using batch size $256$, and a single-cycle cosine scheduler with a base learning rate 1e-4 using the AdamW optimizer~\cite{loshchilovdecoupled}. 

For stage two, we train a Masked Language Model (MLM) to generate images similar to the existing line of works~\cite{chang2022maskgit, mentzerfinite, yu2023magvit, zhaoimage, luo2024open, yu2024language}. We use a cosine mask schedule from MaskGIT~\cite{chang2022maskgit} which demonstrated superior generative performance. The MLM contains 307M parameters, comparable to GPT2-medium capacity. It is trained with the same learning rate and steps as stage one. Following convention, we report FID~\cite{heusel2017gans} and IS~\cite{salimans2016improved} scores for guided generation. More hyperparameter details can be found in Appendix~\ref{app_mlm_hps}.

From Table~\ref{tab:imagenet_1k}, under similar model capacity, FlowVQ $|\mathcal{C}|=16384$ achieves a good balance between reconstruction and generation. It achieves the highest IS score and outperforms all models in terms of gFID with 1.89, except MAGVIT-v2 (1.78), where FlowVQ uses only 6.25\% of its vocabulary size. The baselines include MAR-L-407M, which uses a slightly larger generator and autoregressively generates images in the original continuous latent space without tokenization, as well as models using advanced VQ techniques such as VAR and SQ-VAE-2, which use multi-scale and residual VQ, respectively. Like other VQ methods, FlowVQ is compatible with those techniques.

\begin{figure}[H]
    \centering
    \includegraphics[width=\textwidth]{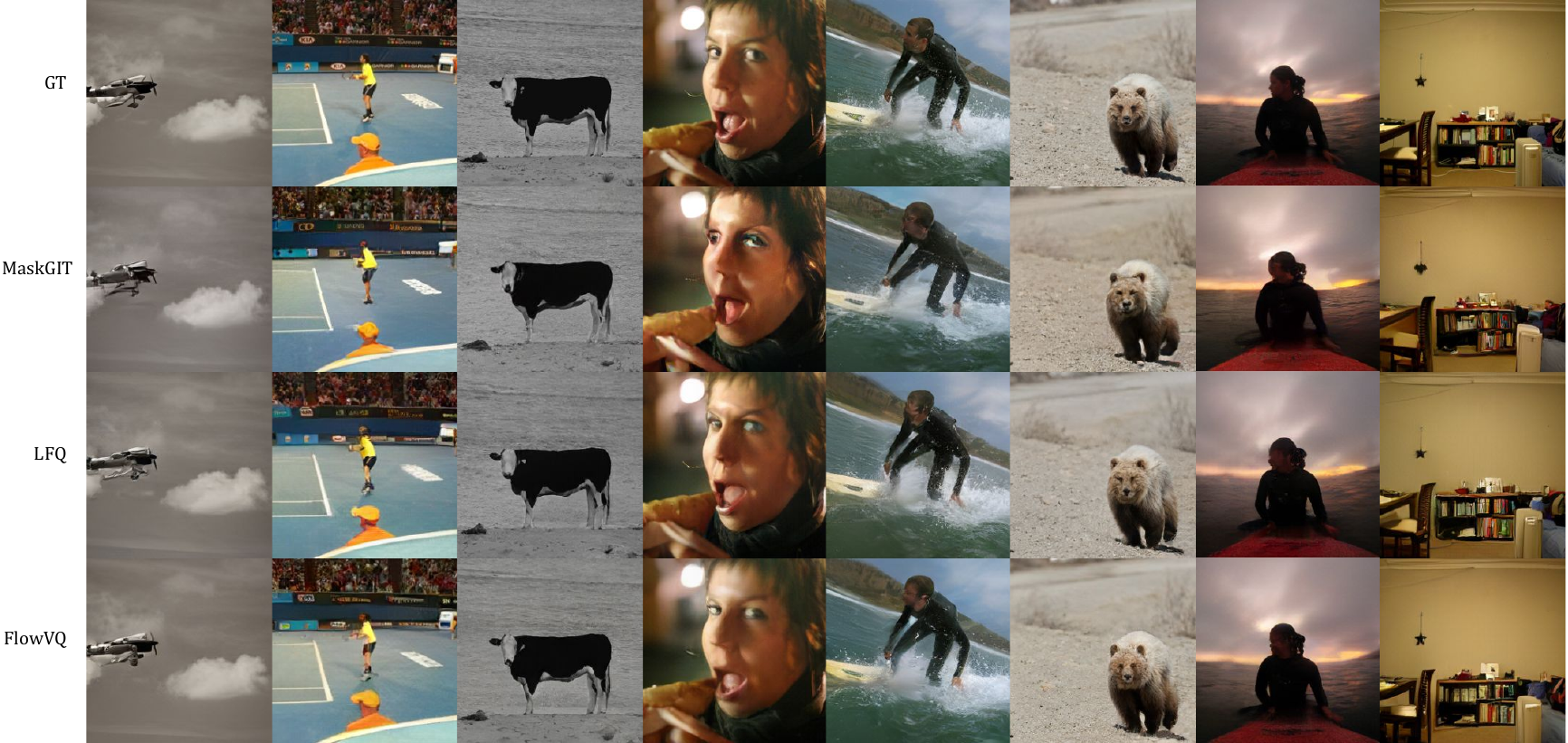}
    \caption{Reconstruction on unseen images from COCO-2017 validation data on ImageNet-1K $256\times256$ trained model.}
    \label{fig: coco_comp}
\end{figure}

For qualitative measure, we plot the generated samples in Figure~\ref{fig: imagenet_gen}. We also validate generalization of the codebook by reconstructing unseen images from COCO-2017 validation set in Figure~\ref{fig: coco_comp}, where we found FlowVQ can produce more faithful and plausible reconstruction over details and particularly faces.

\subsection{High Precision Tokenization}

To examine the potential of an adaptive and fully utilized codebook for FlowVQ as a general tokenizer, we would like to extend the evaluation beyond the image domain. One natural task to consider is high-precision tokenization, especially where structural representation is missing and generation can greatly benefit from the discretized representation learning. Data containing geometric, motion, and spatial state information usually resides in continuous space, and small numerical differences can lead to significantly different physical states or interaction outcomes. Tokenization needs to allocate sufficient resolution to ensure that distinct physical configurations do not collapse into indistinguishable tokens. The current common practice uses k-means initialized VQ-VAE~\cite{pourmirzaei2025gcp, lu2025uni, qiu2024vqcrystal}. We consider two representative tasks here: crystalline material tokenization in Sec~\ref{eval_mattok} and robot action tokenization in Sec~\ref{eval_acttok}.

\subsubsection{Crystalline Material Tokenization} 
\label{eval_mattok}

We evaluate tokenization techniques for crystalline materials based on their ability to faithfully represent crystal structures using an atom-based encoder–decoder transformer. Crystals exhibit periodic atomic arrangements in 3D space and are commonly represented by a unit cell containing a set of atoms. Each atom is described by a tuple $(\mathbf{A}, \mathbf{X}, \mathcal{L})$, where $\mathbf{A}$ is its one-hot atom type vector, $\mathbf{X}$ is its 3D fractional coordinates, and $\mathcal{L}$ is the lattice information. These components are then concatenated into a single vector before being fed into the model.

We evaluate on MP-20~\cite{jain2013commentary} with $|\mathcal{C}| = 8192$ and d = 8. As VQ-VAE and OptVQ benefit from an aligned initialization~\cite{pourmirzaei2025gcp, lu2025uni, qiu2024vqcrystal}, and to isolate the effect of the quantizer, we initialize their codebooks with k-means centroids of the training-set latents from the same autoencoder at epoch W = 500, when FlowVQ quantization begins. All methods also start from the same encoder and decoder weights. For evaluation,  we follow prior work~\cite{joshiall} by comparing each reconstructed material with its ground-truth material using StructureMatcher from PyMatgen~\cite{ong2013python} and reporting the resulting match rate. We do not report codebook utilization rate for this task because MP-20 is a relatively small dataset, and the evaluation distribution does not fully cover the training distribution, which is reflected as only 7,408 of the 8,192 codes derived from the training data k-means clusters were actively used on the evaluation set. This number can further vary with the latent geometry evolving during training. We analyze the number of actively utilized codes instead.

\begin{table}[H]
\centering
\resizebox{\textwidth}{!}{ 
    \begin{tabular}{c|c|c|c|c|c|c|c|c}
        \toprule
         & \shortstack{VQVAE \\ \ }  & \shortstack{LFQ \\ \ } & \shortstack{OptVQ \\ ($\epsilon=10^{*}$) \ } & \shortstack{KM-VQVAE \\ \ } & \shortstack{KM-OptVQ \\ ($\epsilon=10$) } & \shortstack{KM-OptVQ \\ ($\epsilon=1$) } & \shortstack{FlowVQ \\ \ } & \shortstack{FlowVQ \\ (w/ oc)} \\
        \midrule
        Match Rate $\uparrow$ & 47.16 \% & 24.63\% & 10.81\% & 78.38\% & 17.85\% & 53.27\% & \textbf{90.81} \% & 90.66 \% \\
        \bottomrule
        \end{tabular}
}
\vspace{5pt}
\tiny{$^* \epsilon$ in Eq. 8 of the original paper, where $\epsilon=10$ is the open-source configuration across all tasks and used for table~\ref{tab:cifar10} experiments.}
\caption{Match rate between the reconstructed and ground-truth material on MP-20.}
\label{tab:mp_20}
\end{table}
\vspace{-8pt}

FlowVQ significantly outperforms all baseline methods, including k-means initialized techniques. It achieves the highest match rate ($90.81\%$) with 7178 actively utilized codes on average, compared to the second best method KM-VQVAE as $78.38\%$ with the average actively used code number as 3192. LFQ and OptVQ generally perform poorly on this task. For k-means initialized OptVQ, reducing $\epsilon$ from 10 to 1, which strengthens the entropy regularization as its weight scales with $\frac{1}{\epsilon}$, improves the match rate from $17.85\%$ to $53.27\%$ but significantly reduces the number of active codes from 7110 to 1311, consistent with our analysis in Sec~\ref{sec: related_works}. More details on the experiment setting can be found in Appendix~\ref{app_mattok}.

\subsubsection{Robotics Action Tokenization} 
\label{eval_acttok}

Robot actions in foundation models~\cite{kimopenvla, intelligence2025pi_, barreiros2026careful, xue2025reactive} have recently been represented as action chunks~\cite{zhao2023learning, lai2025action}, as fixed-length sequences of actions. Each action is represented as a vector defining the next delta movement. We evaluate on the DROID dataset~\cite{khazatsky2024droid} and compare with common action tokenizers: 1) FAST+~\cite{pertsch2025fast}, which applies the Discrete Cosine Transform (DCT) to each action dimension of a chunk, quantizes the coefficients by scaling and rounding, flattens them from low to high frequency, and compresses the resulting sequence with a pretrained Byte-Pair Encoding (BPE)~\cite{gage1994new, sennrich2016neural} vocabulary, 2) K-means initialized VQ-VAE. Similar to Sec~\ref{eval_mattok}, we use an encoder-decoder transformer with quantization applied in the latent space. $|\mathcal{C}|$ is set to 2048 for fair comparison with existing tokenizers, latent dimension d = 8, and more details can be found in Appendix~\ref{app_acttok}.

\begin{table}[H]
\centering
\begin{tabular}{c|c|c|c}
    \toprule
     & FAST+ & KM-VQVAE  & FlowVQ  \\
    \midrule
    MSE(e-4) $\downarrow$  & 2.40 & 1.88 & \textbf{1.71} \\
    \midrule
    Utilization $\uparrow$  & 97.02\% & 10.89\% & \textbf{100\%} \\
    \bottomrule
\end{tabular}
    \vspace{5pt}
    \caption{Trajectory reconstruction error of 50-step action chunks on DROID dataset.}
    \label{tab:droid}
\end{table}

\vspace{-8pt}
From Table~\ref{tab:droid}, we observe FlowVQ achieves full codebook utilization on validation trajectories throughout the training, and reaches the lowest reconstruction error of $1.71e^{-4}$. KM-VQVAE starts with full utilization, but gradually drops to 10.89\% during training. Both FlowVQ and KM-VQVAE yield a more accurate reconstruction of action compared to FAST+.

\subsection{Ablation Study}

To better understand the contribution of the learned transport and of each codebook correction mechanism, we ablate FlowVQ on CIFAR-10 and MP-20 and report the best model performance in Table~\ref{tab:ablation}. On both tasks, the prior codebook transport can deliver more accurate tokenization compared to baseline methods reflected as lower FID score and higher material match rate (FID 1.49 vs 1.65 in Table~\ref{tab:cifar10}; 88.15\% vs 78.38\% in Table~\ref{tab:mp_20}), and adding self-correction stabilizes the training with less fluctuation observed. Adding online correction on top further improves image tokenization slightly, while performing on par on material tokenization.

\begin{table*}[h] 
    \centering 
    \textbf{(a) Image Tokenization: CIFAR-10} \vspace{5pt} 
    \resizebox{\textwidth}{!}{ 
    \begin{tabular}{lcccccccc} 
        \toprule 
         & \textbf{FID} $\downarrow$ & \textbf{IS} $\uparrow$ & \textbf{SSIM} $\uparrow$ & \textbf{PSNR} $\uparrow$ & \textbf{LPIPS} $\downarrow$ & \textbf{MAE} $\downarrow$ & \textbf{Entropy} $\uparrow$ & \textbf{Util} $\uparrow$ \\ 
        \midrule 
        $\mathcal{C}_0\rightarrow\mathcal{C}_1$ & 1.49 & \textbf{10.79} & 0.856 & 26.11 & 0.086 & 0.038 & 13.92 & 100\% \\ 
        $\mathcal{C}_0\rightarrow\mathcal{C}_1$ + sc & 1.52 & 10.70 & 0.841 & 25.73 & 0.094 & 0.039 & 13.97 & 100\% \\
        $\mathcal{C}_0\rightarrow\mathcal{C}_1$ + sc + oc & \textbf{1.40} & 10.72 & \textbf{0.861} & \textbf{26.50} & \textbf{0.086} & \textbf{0.036} & \textbf{13.97} & 100\% \\ 
        \bottomrule 
    \end{tabular} 
    } 
    \textbf{(b) Crystalline Material Tokenization: MP-20} \vspace{5pt} \\
    \begin{tabular}{lccc}
        \toprule
        \textbf{MP-20 Match Rate} & $\mathcal{C}_0\rightarrow\mathcal{C}_1$  & $\mathcal{C}_0\rightarrow\mathcal{C}_1$ + sc & $\mathcal{C}_0\rightarrow\mathcal{C}_1$ + sc + oc \\
        \midrule
        Match Rate $\uparrow$ & 88.15\% & \textbf{90.81\%} & 90.66\% \\
        \bottomrule
    \end{tabular}
    \caption{Ablation study on FlowVQ components on CIFAR-10 image tokenization and MP-20 material tokenization tasks. With details in Sec~\ref{sec: codebook_optim}, \textit{sc} refers to self-correction mechanism, and \textit{oc} refers to the optional online-correction mechanism.} 
    \label{tab:ablation} 
\end{table*}

\section{Conclusion} 

To summarize, we introduced FlowVQ, a vector quantization technique that reformulates codebook learning in classic VQ-VAE as a quantile-based transport process, where balanced code utilization is governed by the geometry of the latent space rather than a byproduct of auxiliary optimization objectives. By constructing the codebook via deterministic pushforwards from a uniform prior and refining them through self-correction, FlowVQ achieves stable and balanced code usage while avoiding codebook collapse. We conducted experiments across a range of tokenization and generation tasks, demonstrating consistently improved reconstruction and generation quality. We hope the proposed method will serve as a useful tool, especially in fields where features are naturally continuous or hybrid, and benefit from learned representations.

\paragraph{Limitations and Future Work} In addition to the slight increase in training time, FlowVQ primarily focuses on addressing codebook collapse, while another important limitation of VQ-VAE lies in the reliance on the biased STE for backpropagation through discrete quantization. Recent works have shown that this gradient mismatch can be alleviated through simple rotation tricks~\cite{fiftyrestructuring}, but more effective alternatives remain largely unexplored. Another direction worth exploring is to investigate FlowVQ together with recent architectures on flexible-length~\cite{miwa2025one, bachmann2025flextok} or flexible-resolution~\cite{mao2026quadtok} embedding learning, which may enable more adaptive token allocation to regions with higher semantic or structural complexity, leading to further improvements.

\newpage
\paragraph{Acknowledgments} We thank the anonymous reviewers for their constructive feedback and suggestions, which improved the paper. We thank Liu Nian for sharing his code for k-means initialized VQ-VAE for crystalline material tokenization, which enabled an insightful comparison during the development. Zekun Shi is partially supported by the Singapore Economic Development Board through the Industrial Postgraduate Programme. 
\bibliographystyle{unsrtnat}  
\bibliography{reference}  

@article{van2017neural,
  title={Neural discrete representation learning},
  author={Van Den Oord, Aaron and Vinyals, Oriol and others},
  journal={Advances in neural information processing systems},
  volume={30},
  year={2017}
}

@article{razavi2019generating,
  title={Generating diverse high-fidelity images with vq-vae-2},
  author={Razavi, Ali and Van den Oord, Aaron and Vinyals, Oriol},
  journal={Advances in neural information processing systems},
  volume={32},
  year={2019}
}

@inproceedings{bruce2024genie,
  title={Genie: Generative interactive environments},
  author={Bruce, Jake and Dennis, Michael D and Edwards, Ashley and Parker-Holder, Jack and Shi, Yuge and Hughes, Edward and Lai, Matthew and Mavalankar, Aditi and Steigerwald, Richie and Apps, Chris and others},
  booktitle={Forty-first International Conference on Machine Learning},
  year={2024}
}

@inproceedings{rombach2022high,
  title={High-resolution image synthesis with latent diffusion models},
  author={Rombach, Robin and Blattmann, Andreas and Lorenz, Dominik and Esser, Patrick and Ommer, Bj{\"o}rn},
  booktitle={Proceedings of the IEEE/CVF conference on computer vision and pattern recognition},
  pages={10684--10695},
  year={2022}
}

@article{dhariwal2020jukebox,
  title={Jukebox: A generative model for music},
  author={Dhariwal, Prafulla and Jun, Heewoo and Payne, Christine and Kim, Jong Wook and Radford, Alec and Sutskever, Ilya},
  journal={arXiv preprint arXiv:2005.00341},
  year={2020}
}

@inproceedings{zheng2023online,
  title={Online clustered codebook},
  author={Zheng, Chuanxia and Vedaldi, Andrea},
  booktitle={Proceedings of the IEEE/CVF International Conference on Computer Vision},
  pages={22798--22807},
  year={2023}
}

@article{agustsson2017soft,
  title={Soft-to-hard vector quantization for end-to-end learning compressible representations},
  author={Agustsson, Eirikur and Mentzer, Fabian and Tschannen, Michael and Cavigelli, Lukas and Timofte, Radu and Benini, Luca and Gool, Luc V},
  journal={Advances in neural information processing systems},
  volume={30},
  year={2017}
}

@inproceedings{jang2017categorical,
  title={Categorical Reparameterization with Gumbel-Softmax},
  author={Jang, Eric and Gu, Shixiang and Poole, Ben},
  booktitle={International Conference on Learning Representations},
  year={2017}
}

@inproceedings{baevskivq,
  title={vq-wav2vec: Self-Supervised Learning of Discrete Speech Representations},
  author={Baevski, Alexei and Schneider, Steffen and Auli, Michael},
  booktitle={International Conference on Learning Representations},
  year={2020}
}

@article{zhang2024preventing,
  title={Preventing local pitfalls in vector quantization via optimal transport},
  author={Zhang, Borui and Zheng, Wenzhao and Zhou, Jie and Lu, Jiwen},
  journal={arXiv preprint arXiv:2412.15195},
  year={2024}
}

@article{williams2020hierarchical,
  title={Hierarchical quantized autoencoders},
  author={Williams, Will and Ringer, Sam and Ash, Tom and MacLeod, David and Dougherty, Jamie and Hughes, John},
  journal={Advances in Neural Information Processing Systems},
  volume={33},
  pages={4524--4535},
  year={2020}
}

@INPROCEEDINGS{takida2022sq-vae,
    author={Takida, Yuhta and Shibuya, Takashi and Liao, WeiHsiang and Lai, Chieh-Hsin and Ohmura, Junki and Uesaka, Toshimitsu and Murata, Naoki and Takahashi, Shusuke and Kumakura, Toshiyuki and Mitsufuji, Yuki},
    title={{SQ-VAE}: Variational Bayes on Discrete Representation with Self-annealed Stochastic Quantization},
    booktitle={International Conference on Machine Learning},
    year={2022},
    }

@article{takida2024hq,
  title={HQ-VAE: Hierarchical Discrete Representation Learning with Variational Bayes},
  author={Takida, Yuhta and Ikemiya, Yukara and Shibuya, Takashi and Shimada, Kazuki and Choi, Woosung and Lai, Chieh-Hsin and Murata, Naoki and Uesaka, Toshimitsu and Uchida, Kengo and Liao, Wei-Hsiang and others},
  journal={Trans. Mach. Learn. Res.},
  year={2024}
}

@article{zhu2024scaling,
  title={Scaling the codebook size of VQ-GAN to 100,000 with a utilization rate of 99\%},
  author={Zhu, Lei and Wei, Fangyun and Lu, Yanye and Chen, Dong},
  journal={Advances in Neural Information Processing Systems},
  volume={37},
  pages={12612--12635},
  year={2024}
}

@article{chang2026scalable,
  title={Scalable training for vector-quantized networks with 100\% codebook utilization},
  author={Chang, Yifan and Qin, Jie and Qiao, Limeng and Wang, Xiaofeng and Zhu, Zheng and Ma, Lin and Wang, Xingang},
  journal={International Conference on Learning Representations},
  year={2026}
}

@inproceedings{chang2022maskgit,
  title={Maskgit: Masked generative image transformer},
  author={Chang, Huiwen and Zhang, Han and Jiang, Lu and Liu, Ce and Freeman, William T},
  booktitle={Proceedings of the IEEE/CVF conference on computer vision and pattern recognition},
  pages={11315--11325},
  year={2022}
}

@inproceedings{mentzerfinite,
  title={Finite Scalar Quantization: VQ-VAE Made Simple},
  author={Mentzer, Fabian and Minnen, David and Agustsson, Eirikur and Tschannen, Michael},
  booktitle={The Twelfth International Conference on Learning Representations},
  year={2024}
}

@inproceedings{yu2024language,
  title={Language Model Beats Diffusion-Tokenizer is key to visual generation},
  author={Yu, Lijun and Lezama, Jos{\'e} and Gundavarapu, Nitesh Bharadwaj and Versari, Luca and Sohn, Kihyuk and Minnen, David and Cheng, Yong and Gupta, Agrim and Gu, Xiuye and Hauptmann, Alexander G and others},
  booktitle={ICLR},
  year={2024}
}

@article{webermaskbit,
  title={MaskBit: Embedding-free Image Generation via Bit Tokens},
  author={Weber, Mark and Yu, Lijun and Yu, Qihang and Deng, Xueqing and Shen, Xiaohui and Cremers, Daniel and Chen, Liang-Chieh},
  journal={Transactions on Machine Learning Research},
  year={2024}
}

@inproceedings{zhaoimage,
  title={Image and Video Tokenization with Binary Spherical Quantization},
  author={Zhao, Yue and Xiong, Yuanjun and Kraehenbuehl, Philipp},
  booktitle={The Thirteenth International Conference on Learning Representations},
  year={2025}
}

@article{song2020score,
  title={Score-based generative modeling through stochastic differential equations},
  author={Song, Yang and Sohl-Dickstein, Jascha and Kingma, Diederik P and Kumar, Abhishek and Ermon, Stefano and Poole, Ben},
  journal={arXiv preprint arXiv:2011.13456},
  year={2020}
}

@article{maoutsa2020interacting,
  title={Interacting particle solutions of fokker--planck equations through gradient--log--density estimation},
  author={Maoutsa, Dimitra and Reich, Sebastian and Opper, Manfred},
  journal={Entropy},
  volume={22},
  number={8},
  pages={802},
  year={2020},
  publisher={MDPI}
}

@article{lipman2022flow,
  title={Flow matching for generative modeling},
  author={Lipman, Yaron and Chen, Ricky TQ and Ben-Hamu, Heli and Nickel, Maximilian and Le, Matt},
  journal={arXiv preprint arXiv:2210.02747},
  year={2022}
}

@article{van2016conditional,
  title={Conditional image generation with pixelcnn decoders},
  author={Van den Oord, Aaron and Kalchbrenner, Nal and Espeholt, Lasse and Vinyals, Oriol and Graves, Alex and others},
  journal={Advances in neural information processing systems},
  volume={29},
  year={2016}
}

@article{tian2024visual,
  title={Visual autoregressive modeling: Scalable image generation via next-scale prediction},
  author={Tian, Keyu and Jiang, Yi and Yuan, Zehuan and Peng, Bingyue and Wang, Liwei},
  journal={Advances in neural information processing systems},
  volume={37},
  pages={84839--84865},
  year={2024}
}

@article{dieleman2018challenge,
  title={The challenge of realistic music generation: modelling raw audio at scale},
  author={Dieleman, Sander and Van Den Oord, Aaron and Simonyan, Karen},
  journal={Advances in neural information processing systems},
  volume={31},
  year={2018}
}

@article{yan2021videogpt,
  title={Videogpt: Video generation using vq-vae and transformers},
  author={Yan, Wilson and Zhang, Yunzhi and Abbeel, Pieter and Srinivas, Aravind},
  journal={arXiv preprint arXiv:2104.10157},
  year={2021}
}

@article{pourmirzaei2025gcp,
  title={GCP-VQVAE: A Geometry-Complete Language for Protein 3D Structure},
  author={Pourmirzaei, Mahdi and Morehead, Alex and Esmaili, Farzaneh and Ren, Jarett and Pourmirzaei, Mohammadreza and Xu, Dong},
  journal={bioRxiv},
  pages={2025--10},
  year={2025},
  publisher={Cold Spring Harbor Laboratory}
}

@article{lu2025uni,
  title={Uni-3dar: Unified 3d generation and understanding via autoregression on compressed spatial tokens},
  author={Lu, Shuqi and Lin, Haowei and Yao, Lin and Gao, Zhifeng and Ji, Xiaohong and Zhang, Linfeng and Ke, Guolin and others},
  journal={arXiv preprint arXiv:2503.16278},
  year={2025}
}

@article{qiu2024vqcrystal,
  title={VQCrystal: Leveraging Vector Quantization for Discovery of Stable Crystal Structures},
  author={Qiu, ZiJie and Jin, Luozhijie and Du, Zijian and Chen, Hongyu and Cen, Yan and Sun, Siqi and Mei, Yongfeng and Zhang, Hao},
  journal={arXiv preprint arXiv:2409.06191},
  year={2024}
}

@inproceedings{lee2022autoregressive,
  title={Autoregressive image generation using residual quantization},
  author={Lee, Doyup and Kim, Chiheon and Kim, Saehoon and Cho, Minsu and Han, Wook-Shin},
  booktitle={Proceedings of the IEEE/CVF conference on computer vision and pattern recognition},
  pages={11523--11532},
  year={2022}
}

@inproceedings{esser2021taming,
  title={Taming transformers for high-resolution image synthesis},
  author={Esser, Patrick and Rombach, Robin and Ommer, Bjorn},
  booktitle={Proceedings of the IEEE/CVF conference on computer vision and pattern recognition},
  pages={12873--12883},
  year={2021}
}

@inproceedings{deng2009imagenet,
  title={Imagenet: A large-scale hierarchical image database},
  author={Deng, Jia and Dong, Wei and Socher, Richard and Li, Li-Jia and Li, Kai and Fei-Fei, Li},
  booktitle={2009 IEEE conference on computer vision and pattern recognition},
  pages={248--255},
  year={2009},
  organization={Ieee}
}

@article{krizhevsky2009learning,
  title={Learning multiple layers of features from tiny images},
  author={Krizhevsky, Alex and Hinton, Geoffrey and others},
  year={2009},
  publisher={Toronto, ON, Canada}
}

@article{li2024autoregressive,
  title={Autoregressive image generation without vector quantization},
  author={Li, Tianhong and Tian, Yonglong and Li, He and Deng, Mingyang and He, Kaiming},
  journal={Advances in Neural Information Processing Systems},
  volume={37},
  pages={56424--56445},
  year={2024}
}

@article{pertsch2025fast,
  title={Fast: Efficient action tokenization for vision-language-action models},
  author={Pertsch, Karl and Stachowicz, Kyle and Ichter, Brian and Driess, Danny and Nair, Suraj and Vuong, Quan and Mees, Oier and Finn, Chelsea and Levine, Sergey},
  journal={arXiv preprint arXiv:2501.09747},
  year={2025}
}

@article{intelligence2025pi_,
  title={$\pi_{0.5}$: a Vision-Language-Action Model with Open-World Generalization},
  author={Intelligence, Physical and Black, Kevin and Brown, Noah and Darpinian, James and Dhabalia, Karan and Driess, Danny and Esmail, Adnan and Equi, Michael and Finn, Chelsea and Fusai, Niccolo and others},
  journal={arXiv preprint arXiv:2504.16054},
  year={2025}
}

@inproceedings{joshiall,
  title={All-atom Diffusion Transformers: Unified generative modelling of molecules and materials},
  author={Joshi, Chaitanya K and Fu, Xiang and Liao, Yi-Lun and Gharakhanyan, Vahe and Miller, Benjamin Kurt and Sriram, Anuroop and Ulissi, Zachary Ward},
  booktitle={Forty-second International Conference on Machine Learning},
  year={2025}
}

@article{ong2013python,
  title={Python Materials Genomics (pymatgen): A robust, open-source python library for materials analysis},
  author={Ong, Shyue Ping and Richards, William Davidson and Jain, Anubhav and Hautier, Geoffroy and Kocher, Michael and Cholia, Shreyas and Gunter, Dan and Chevrier, Vincent L and Persson, Kristin A and Ceder, Gerbrand},
  journal={Computational Materials Science},
  volume={68},
  pages={314--319},
  year={2013},
  publisher={Elsevier}
}

@article{jain2013commentary,
  title={Commentary: The Materials Project: A materials genome approach to accelerating materials innovation},
  author={Jain, Anubhav and Ong, Shyue Ping and Hautier, Geoffroy and Chen, Wei and Richards, William Davidson and Dacek, Stephen and Cholia, Shreyas and Gunter, Dan and Skinner, David and Ceder, Gerbrand and others},
  journal={APL materials},
  volume={1},
  number={1},
  year={2013},
  publisher={AIP Publishing}
}

@article{douze2025faiss,
  title={The faiss library},
  author={Douze, Matthijs and Guzhva, Alexandr and Deng, Chengqi and Johnson, Jeff and Szilvasy, Gergely and Mazar{\'e}, Pierre-Emmanuel and Lomeli, Maria and Hosseini, Lucas and J{\'e}gou, Herv{\'e}},
  journal={IEEE Transactions on Big Data},
  year={2025},
  publisher={IEEE}
}

@article{zhao2023learning,
  title={Learning fine-grained bimanual manipulation with low-cost hardware},
  author={Zhao, Tony Z and Kumar, Vikash and Levine, Sergey and Finn, Chelsea},
  journal={arXiv preprint arXiv:2304.13705},
  year={2023}
}

@article{khazatsky2024droid,
  title={Droid: A large-scale in-the-wild robot manipulation dataset},
  author={Khazatsky, Alexander and Pertsch, Karl and Nair, Suraj and Balakrishna, Ashwin and Dasari, Sudeep and Karamcheti, Siddharth and Nasiriany, Soroush and Srirama, Mohan Kumar and Chen, Lawrence Yunliang and Ellis, Kirsty and others},
  journal={arXiv preprint arXiv:2403.12945},
  year={2024}
}

@article{krause2010discriminative,
  title={Discriminative clustering by regularized information maximization},
  author={Krause, Andreas and Perona, Pietro and Gomes, Ryan},
  journal={Advances in neural information processing systems},
  volume={23},
  year={2010}
}

@inproceedings{jansen2020coincidence,
  title={Coincidence, categorization, and consolidation: Learning to recognize sounds with minimal supervision},
  author={Jansen, Aren and Ellis, Daniel PW and Hershey, Shawn and Moore, R Channing and Plakal, Manoj and Popat, Ashok C and Saurous, Rif A},
  booktitle={ICASSP 2020-2020 IEEE International Conference on Acoustics, Speech and Signal Processing (ICASSP)},
  pages={121--125},
  year={2020},
  organization={IEEE}
}

@article{heusel2017gans,
  title={Gans trained by a two time-scale update rule converge to a local nash equilibrium},
  author={Heusel, Martin and Ramsauer, Hubert and Unterthiner, Thomas and Nessler, Bernhard and Hochreiter, Sepp},
  journal={Advances in neural information processing systems},
  volume={30},
  year={2017}
}

@article{salimans2016improved,
  title={Improved techniques for training gans},
  author={Salimans, Tim and Goodfellow, Ian and Zaremba, Wojciech and Cheung, Vicki and Radford, Alec and Chen, Xi},
  journal={Advances in neural information processing systems},
  volume={29},
  year={2016}
}

@inproceedings{loshchilovdecoupled,
  title={Decoupled Weight Decay Regularization},
  author={Loshchilov, Ilya and Hutter, Frank},
  booktitle={International Conference on Learning Representations},
  year={2019}
}

@article{sun2024autoregressive,
  title={Autoregressive model beats diffusion: Llama for scalable image generation},
  author={Sun, Peize and Jiang, Yi and Chen, Shoufa and Zhang, Shilong and Peng, Bingyue and Luo, Ping and Yuan, Zehuan},
  journal={arXiv preprint arXiv:2406.06525},
  year={2024}
}

@article{luo2024open,
  title={Open-magvit2: An open-source project toward democratizing auto-regressive visual generation},
  author={Luo, Zhuoyan and Shi, Fengyuan and Ge, Yixiao and Yang, Yujiu and Wang, Limin and Shan, Ying},
  journal={arXiv preprint arXiv:2409.04410},
  year={2024}
}

@inproceedings{yu2023magvit,
  title={Magvit: Masked generative video transformer},
  author={Yu, Lijun and Cheng, Yong and Sohn, Kihyuk and Lezama, Jos{\'e} and Zhang, Han and Chang, Huiwen and Hauptmann, Alexander G and Yang, Ming-Hsuan and Hao, Yuan and Essa, Irfan and others},
  booktitle={Proceedings of the IEEE/CVF Conference on Computer Vision and Pattern Recognition},
  pages={10459--10469},
  year={2023}
}

@article{c,
  title={Language models are unsupervised multitask learners},
  author={Radford, Alec and Wu, Jeffrey and Child, Rewon and Luan, David and Amodei, Dario and Sutskever, Ilya and others},
  journal={OpenAI blog},
  volume={1},
  number={8},
  pages={9},
  year={2019}
}

@article{lai2025action,
  title={Action chunking as conditional policy compression},
  author={Lai, Lucy and Huang, Ann ZX and Gershman, Samuel J},
  journal={Cognition},
  volume={264},
  pages={106201},
  year={2025},
  publisher={Elsevier}
}

@article{barreiros2026careful,
  title={A careful examination of large behavior models for multitask dexterous manipulation},
  author={Barreiros, Jose and Beaulieu, Andrew and Bhat, Aditya and Cory, Rick and Cousineau, Eric and Dai, Hongkai and Fang, Ching-Hsin and Hashimoto, Kunimatsu and Irshad, Muhammad Zubair and Itkina, Masha and others},
  journal={Science Robotics},
  volume={11},
  number={113},
  pages={eaea6201},
  year={2026},
  publisher={American Association for the Advancement of Science}
}

@inproceedings{kimopenvla,
  title={OpenVLA: An Open-Source Vision-Language-Action Model},
  author={Kim, Moo Jin and Pertsch, Karl and Karamcheti, Siddharth and Xiao, Ted and Balakrishna, Ashwin and Nair, Suraj and Rafailov, Rafael and Foster, Ethan P and Sanketi, Pannag R and Vuong, Quan and others},
  booktitle={8th Annual Conference on Robot Learning},
  year={2024}
}

@inproceedings{xue2025reactive,
  title     = {Reactive Diffusion Policy: Slow-Fast Visual-Tactile Policy Learning for Contact-Rich Manipulation},
  author    = {Xue, Han and Ren, Jieji and Chen, Wendi and Zhang, Gu and Fang, Yuan and Gu, Guoying and Xu, Huazhe and Lu, Cewu},
  booktitle = {Proceedings of Robotics: Science and Systems (RSS)},
  year      = {2025}
}

@article{zeni2025generative,
  title={A generative model for inorganic materials design},
  author={Zeni, Claudio and Pinsler, Robert and Z{\"u}gner, Daniel and Fowler, Andrew and Horton, Matthew and Fu, Xiang and Wang, Zilong and Shysheya, Aliaksandra and Crabb{\'e}, Jonathan and Ueda, Shoko and others},
  journal={Nature},
  volume={639},
  number={8055},
  pages={624--632},
  year={2025},
  publisher={Nature Publishing Group UK London}
}

@article{zheng2022movq,
  title={Movq: Modulating quantized vectors for high-fidelity image generation},
  author={Zheng, Chuanxia and Vuong, Tung-Long and Cai, Jianfei and Phung, Dinh},
  journal={Advances in Neural Information Processing Systems},
  volume={35},
  pages={23412--23425},
  year={2022}
}

@article{jiang2026differentiable,
  title={Differentiable Vector Quantization for Rate-Distortion Optimization of Generative Image Compression},
  author={Jiang, Shiyin and Long, Wei and Han, Minghao and Chen, Zhenghao and Zhu, Ce and Gu, Shuhang},
  journal={arXiv preprint arXiv:2604.10546},
  year={2026}
}

@article{miwa2025one,
  title={One-d-piece: Image tokenizer meets quality-controllable compression},
  author={Miwa, Keita and Sasaki, Kento and Arai, Hidehisa and Takahashi, Tsubasa and Yamaguchi, Yu},
  journal={arXiv preprint arXiv:2501.10064},
  year={2025}
}

@inproceedings{bachmann2025flextok,
  title={Flextok: Resampling images into 1d token sequences of flexible length},
  author={Bachmann, Roman and Allardice, Jesse and Mizrahi, David and Fini, Enrico and Kar, O{\u{g}}uzhan Fatih and Amirloo, Elmira and El-Nouby, Alaaeldin and Zamir, Amir and Dehghan, Afshin},
  booktitle={Forty-second International Conference on Machine Learning},
  year={2025}
}

@inproceedings{fiftyrestructuring,
  title={Restructuring Vector Quantization with the Rotation Trick},
  author={Fifty, Christopher and Junkins, Ronald Guenther and Duan, Dennis and Iyengar, Aniketh and Liu, Jerry Weihong and Amid, Ehsan and Thrun, Sebastian and Re, Christopher},
  booktitle={The Thirteenth International Conference on Learning Representations},
  year={2025}
}

@article{mao2026quadtok,
  title={QuadTok: Quadtree Visual Tokenizer for Autoregressive Image Generation},
  author={Mao, Yucheng and Chen, Zeyuan and Shan, Xiaojun and Zhang, Xiang and Srivastava, Divyansh and Li, Bingnan and Tu, Zhuowen},
  journal={arXiv preprint arXiv:2610.10497},
  year={2026}
}

@article{liu2026composable,
  title={Composable Crystals: Controllable Materials Discovery via Concept Learning},
  author={Liu, Nian and Zeng, Yuwei and Kubo, Ryoji and Kazeev, Nikita and Dale, Stephen Gregory and Maevskiy, Artem and Huang, Pengru and Laurent, Thomas and Novoselov, Kostya S and Bresson, Xavier},
  journal={arXiv preprint arXiv:2605.14769},
  year={2026}
}

@article{gage1994new,
  title={A new algorithm for data compression},
  author={Gage, Philip},
  journal={C Users Journal},
  volume={12},
  number={2},
  pages={23--38},
  year={1994},
  publisher={McPherson, KS: R \& D Publications, c1987-1994.}
}

@inproceedings{sennrich2016neural,
  title={Neural machine translation of rare words with subword units},
  author={Sennrich, Rico and Haddow, Barry and Birch, Alexandra},
  booktitle={Proceedings of the 54th annual meeting of the association for computational linguistics (volume 1: long papers)},
  pages={1715--1725},
  year={2016}
}

\clearpage
\appendix

{\LARGE \textbf{Appendix}}

\section{Algorithm}
\label{app_algo}
Here we now present the pseudocode of the complete training scheme for FlowVQ in Algorithm~\ref{alg:flow-vq}.
\begin{algorithm}[H]
    \caption{Flow-based Vector Quantization (FlowVQ)}
    \label{alg:flow-vq}
    \begin{algorithmic}[1]
        \REQUIRE Dataset $\mathcal{D}$, Encoder $E_\theta$, Decoder $D_\theta$, Quantizer $\psi_\gamma$ \\ 
        \REQUIRE $T$ - total steps, $K$ - codebook size, $d$ - codebook dimension, $N$ - dense sample ratio\\
        \quad \quad \ \ $W$ - starting steps, $M$ - codebook update frequency , $\beta$ - $\mathcal{L}_{\text{commitment}}$ coefficient\\
        \STATE Initialize prior codebook and dense samples: \[\{\mathbf{C}_0^i\}_{i=1}^{K} = \mathcal{F}^{-1}(\textit{Sobol}(K, d)), \quad\{\mathbf{S}_0^i\}_{i=1}^{N K} = \mathcal{F}^{-1}(\textit{Sobol}(N K, d))\]
        
        \FOR{$i = 1$ \textbf{to} T}
            \IF{$i < W$}
                \STATE $\mathcal{L}_\text{recon}(\theta) = \| D_\theta(E_\theta(x)) - x \|_2^2$ 
                \STATE $\theta \gets \theta - \eta \nabla_{\theta}\mathcal{L}_\text{recon}$
            \ELSE[\textit{Train tokenization with FlowVQ}]
                \IF[\textit{Initialize flow $\psi_\gamma$ at step W}]{$i = W$} 
                    \WHILE{not converged} 
                        \STATE $\mathcal{L}_\text{FM}(\gamma) = \mathbb{E}\big[\| u_t^{\gamma}(z) - u_t(z | E_\theta(x), z_0) \|_2^2 \big]$ for $x \in \mathcal{D}$
                        \STATE $\gamma \gets \gamma - \eta \nabla_{\gamma}\mathcal{L}_\text{FM}$
                    \ENDWHILE
                \ENDIF
                \IF[\textit{Update $\mathcal{C}_1$ periodically afterwards}]{$(i-W) \bmod M = 0$} 
                    \STATE $\mathcal{C}_1 = \psi_1^{\gamma}(\mathcal{C}_0), \mathcal{S}_1 = \psi_1^{\gamma}(\mathcal{S}_0)$
                    \STATE $\mathcal{C}_1 \gets \arg\min_{\mathcal{C}_1} \sum_{j=1}^K \sum_{s_1^i \in Q_j} \| s_1^i - \mathbf{C}_1^j \|_2^2$
                \ENDIF
                \STATE $z_e = E_{\theta}(x)$
                \STATE $z_q = \text{VQ}(z_e, \mathcal{C}_1)$
                \STATE $\mathcal{L}_\text{FlowVQ}(\theta, \gamma) = \| D_{\theta}(\text{STE}(z_q)) - x \|_2^2 + \beta \| z_e - \text{sg}(z_q) \|_2^2 + \|u^\gamma_t(z) - u_t(z\mid z_e, z_0) \|_2^2$
                \STATE $(\theta, \gamma) \gets (\theta, \gamma) - \eta \nabla_{(\theta, \gamma)}\mathcal{L}_\text{FlowVQ}$

            \ENDIF
        \ENDFOR
    \end{algorithmic}
\end{algorithm}

\section{Training Time Comparison}
\label{app_train_time}

 Here we report the mean training time per 1000 steps as one codebook update cycle using $|\mathcal{C}|=16384$. 

 For training time, the relatively higher variance of FlowVQ is due to the usage of adaptive ODE solver Dopri45, where we did not investigate alternative solvers in this work. In particular, fixed-step Euler integration may offer further speedups without significantly compromising performance, a possibility that remains to be explored.

\begin{table}[H]
\centering
\resizebox{0.7\textwidth}{!}{    
    \begin{tabular}{c|cccc}
        \toprule
        & VQ-GAN & LFQ & OptVQ & FlowVQ w/ sc \\
        \midrule
         t (s)$\downarrow$ & $535.2 \pm 0.4$ & $578.2 \pm0.4$ & $596.4\pm0.5$ & $618.0\pm13.5$\\
        \bottomrule
    \end{tabular}
    }
    \vspace{10pt}
\caption{Runtime per 1,000 steps (one codebook update cycle) on a single NVIDIA A5000 GPU, with no other processes running and $|\mathcal{C}|=16384$.}
\label{tab:time_analaysis}
\end{table}

\clearpage
\section{Image Tokenization}

\subsection{Image Reconstruction}
\label{app_recon_hps}

The hyper-parameters for image tokenization tasks for both CIFAR-10 and ImageNet-1K $256\times256$ are listed in Table~\ref{tab:image_recon_hps}:

\begin{table}[H]
\centering
\resizebox{\textwidth}{!}{    
\begin{tabular}{c|cc}
        \toprule
         \textbf{Config} & \textbf{CIFAR-10 Reconstruction} &  \textbf{ImageNet-1k Reconstruction} \\
        \midrule
        input size & $32\times 32 \times3$ & $256 \times 256 \times3$ \\ 
        latent size & $8 \times 8 \times d$ & $16 \times 16 \times d$ \\
        batch size & 128 & 256 \\
        \midrule
        \textbf{Data Augmentation} & False & True\\
        \midrule
        random cropping &    & [0.8, 1.0] \\
        random aspect ratio &   & [$\dfrac{3}{4}$, $\dfrac{4}{3}$] \\
        resize to input size &   & True \\
        random horizontal flip &   & True \\
        \midrule
        \textbf{Optimizer} &  \multicolumn{2}{c}{AdamW}\\
        \midrule
        base lr &  \multicolumn{2}{c}{$1e^{-4}$} \\
        beta & \multicolumn{2}{c}{[0.9, 0.999]} \\
        weight decay & \multicolumn{2}{c}{$1e^{-4}$} \\
        warmup step & 0 & 5000 \\
        scheduler &  \multicolumn{2}{c}{Cosine} \\
        gradient clipping norm &  \multicolumn{2}{c}{1.0} \\
        total steps & $|C|$ = 1024, 10240, 16384 - 165K, 165K, 325K &  1.35M\\
        \midrule
        \textbf{Model} & & \\
        \midrule
        hidden dimension & \multicolumn{2}{c}{128} \\
        channel multiplier & [1, 2, 4] & [1, 1, 2, 2, 4] \\
        residual blocks & 2 & 2 \\
        \midrule
        \textbf{Loss Coefficients} & & \\
        \midrule
        discriminator start step &    \multicolumn{2}{c}{20k} \\
        $\beta$ & \multicolumn{2}{c}{0.25} \\
        $\mathcal{L}_{reconstruction}$ weight & \multicolumn{2}{c}{4.0} \\
        $\mathcal{L}_{discriminator}$  & \multicolumn{2}{c}{0.02} \\
        $\mathcal{L}_{perceptual}$  & \multicolumn{2}{c}{0.1} \\
        $\mathcal{L}_{LeCam}$  & \multicolumn{2}{c}{0.001} \\
        $\mathcal{L}_{Entropy}$  & \multicolumn{2}{c}{0.02} \\
        $\mathcal{L}_{fm}$  & \multicolumn{2}{c}{1} \\
        GPU &  A5000 & A100 \\
        
        \bottomrule
    \end{tabular}
    }
\caption{Hyperparameters for image reconstruction.}
\label{tab:image_recon_hps}
\end{table}

\clearpage
\subsection{Masked Language Model}
\label{app_mlm_hps}
The hyper-parameters for image generation using masked language model is listed here for ImageNet-1K $256\times256$ are listed in Table~\ref{tab:image_mlm_hps} below:

\begin{table}[H]
\centering
    \begin{tabular}{c|c}
        \toprule
         \textbf{Config} &  \\
        \midrule
        \#Tokens & $16\times 16$ \\ 
        batch size & 1024 \\
        \textbf{Data Augmentation} &\\
        random cropping& [0.8, 1.0] \\
        resize to input size  & True \\
        random horizontal flip  & True \\
        \textbf{Optimizer} & AdamW\\
        base lr &  $1e^{-4}$ \\
        beta & [0.9, 0.96] \\
        weight decay & 0.045 \\
        warmup step & 5000 \\
        scheduler &  Cosine \\
        gradient clipping norm &  1.0 \\
        total steps &  1.35M\\
        \textbf{Model} &  \\
        hidden dimension & 4096 \\
        attention heads & 16 \\
        depth &  24 \\
        dropout & 0.1 \\
        label dropout & 0.1 \\
        class label smoothing & 0.1 \\ 
        mask schedule & arccos \\
        \textbf{classifier free sampling}  & \\
        temperature & 9.0 \\
        guidance & 7.6 \\
        scale power & 2.5 \\
        GPU &  A100 \\
        
        \bottomrule
    \end{tabular}

\caption{Hyperparameters for masked language model on image generation.}
\label{tab:image_mlm_hps}
\end{table}
\vspace{-15pt}

\begin{figure}[H]
    \centering
    \includegraphics[width=0.6\textwidth]{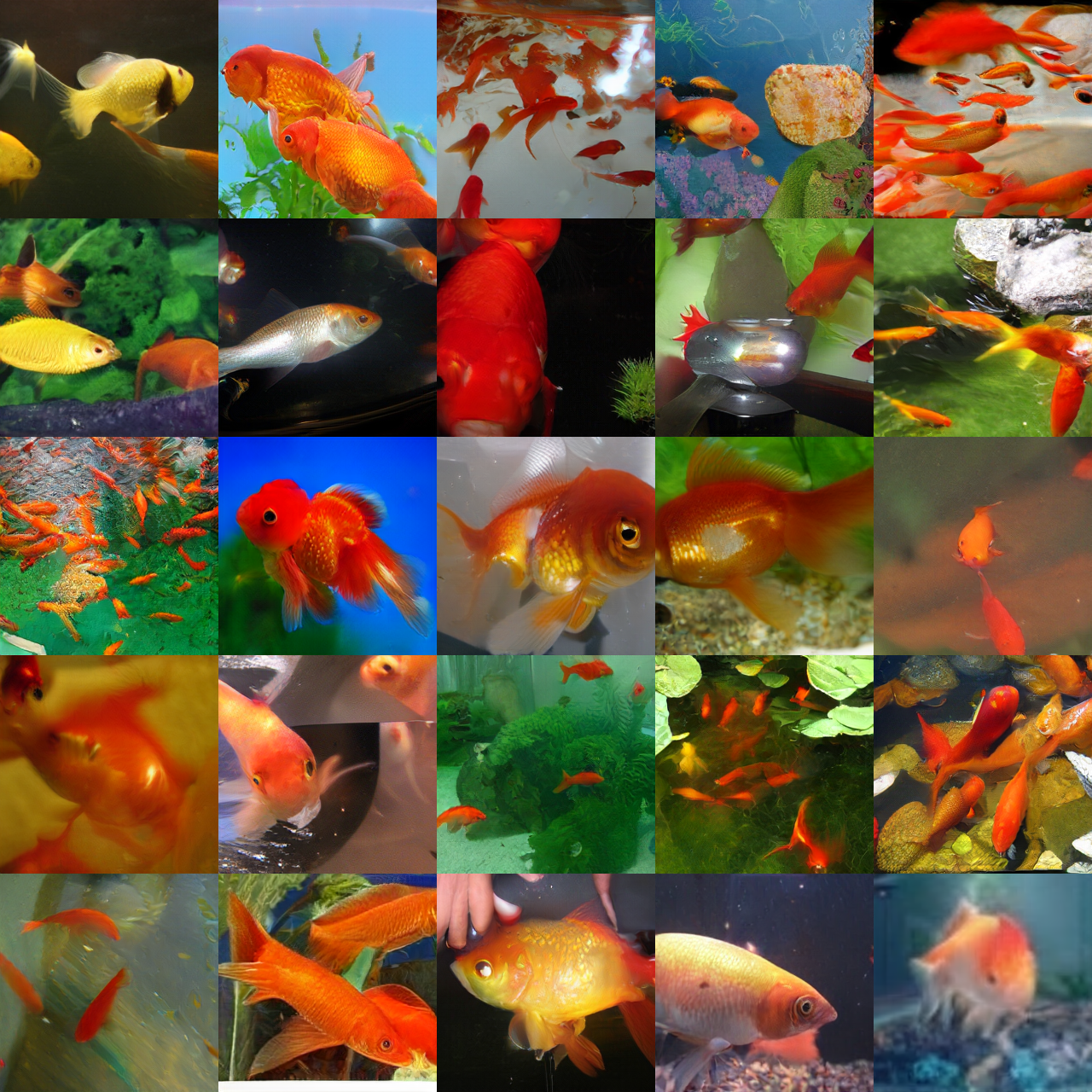}
    \label{fig: gen_c1}
    \vspace{-6pt}
    \caption{goldfish, Carassius auratus}
\end{figure}

\subsection{More Generated Samples}
\label{app_gen_samples}

In this section, we present more samples using classifier free guidance. The model is given the same label in the batch size 25. We reshape the output to [$5\times256$, $5\times256$] and plot here over different classes:

\begin{figure}[H]
    \centering
    \includegraphics[width=0.65\textwidth]{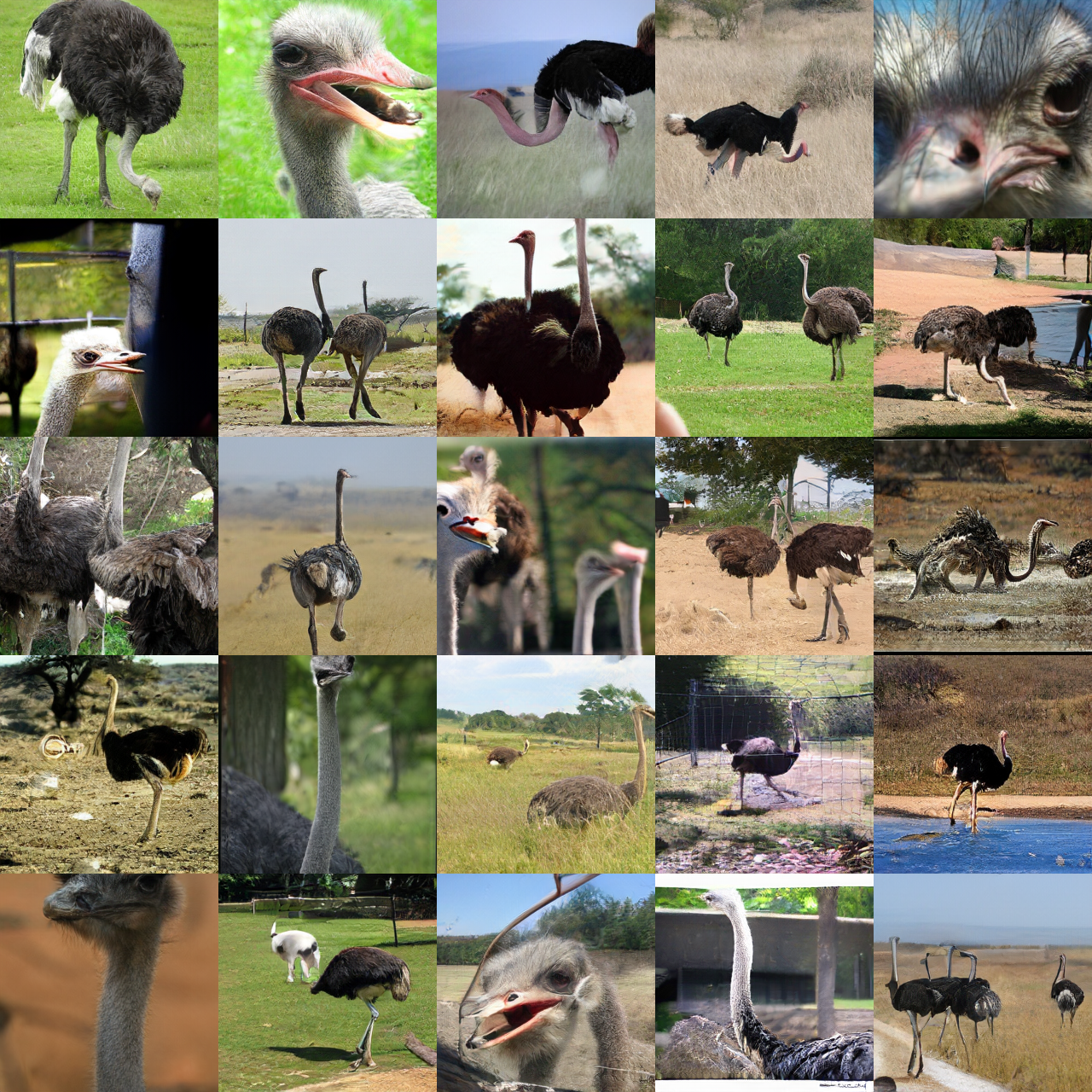}
    \label{fig: gen_c9}
    \caption{ostrich, Struthio camelus}
\end{figure} 

\begin{figure}[H]
    \centering
    \includegraphics[width=0.65\textwidth]{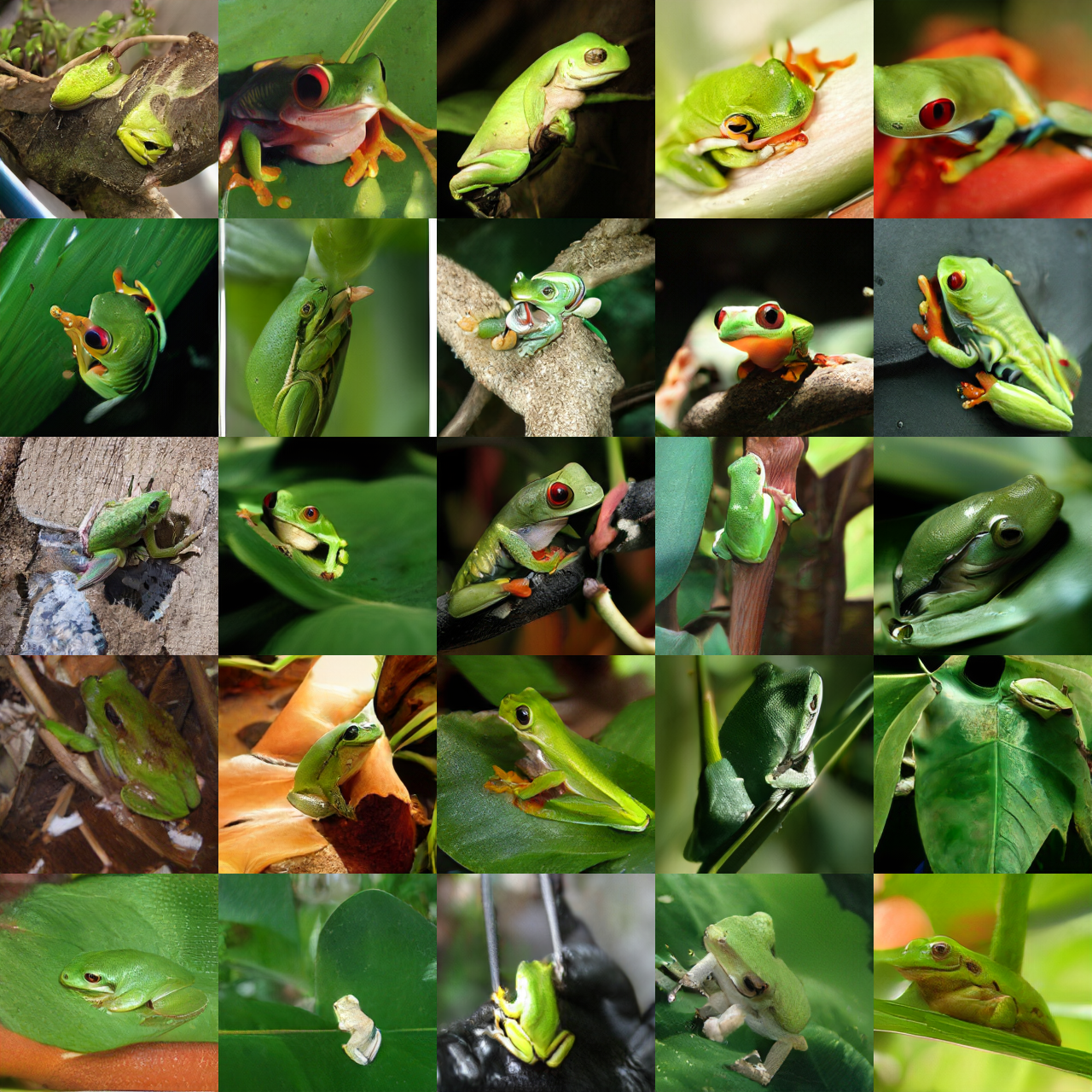}
    \label{fig: gen_c31}
    \caption{tree frog, tree-frog}
\end{figure}

\begin{figure}[H]
    \centering
    \includegraphics[width=0.65\textwidth]{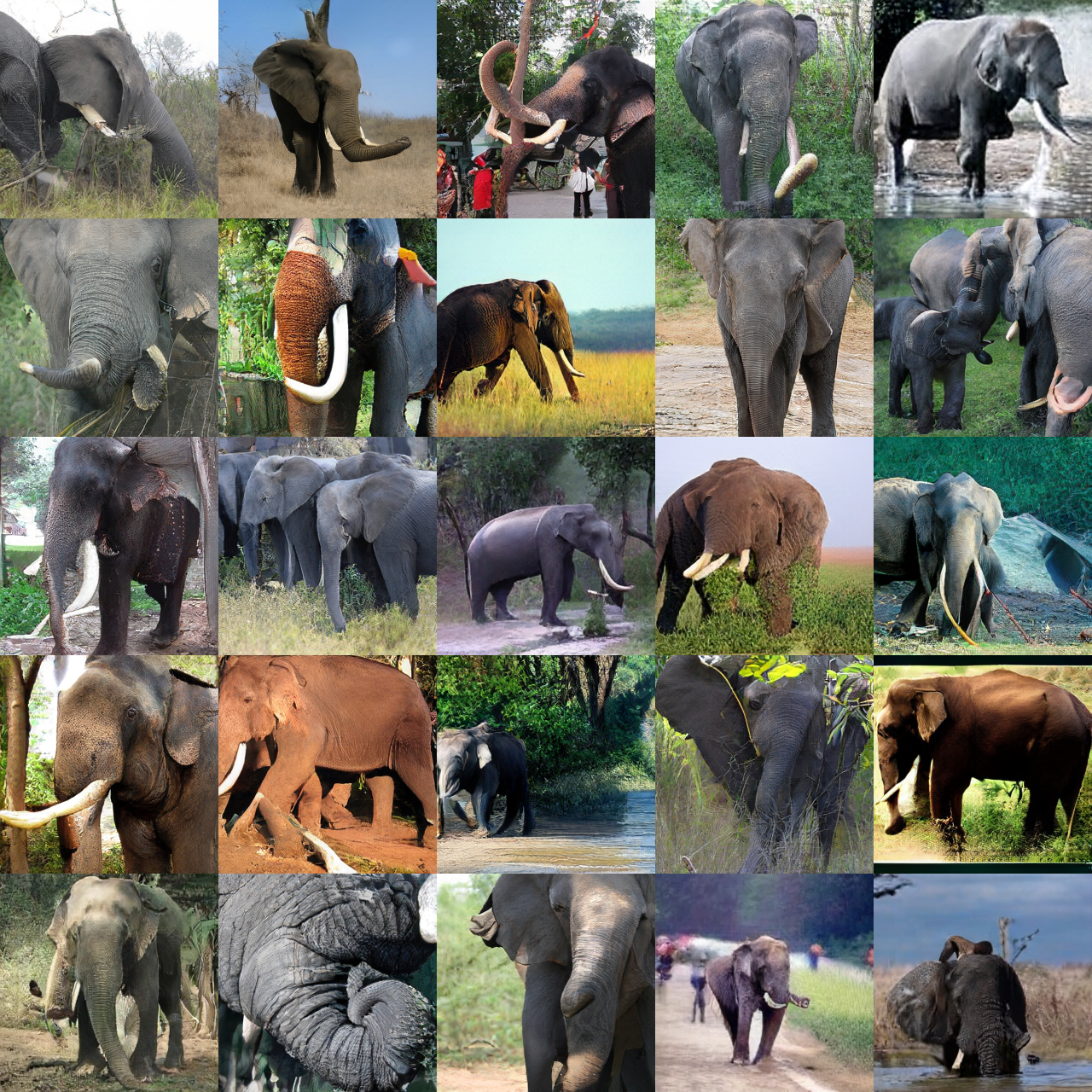}
    \label{fig: gen_c101}
    \caption{tusker}
\end{figure}

\begin{figure}[H]
    \centering
    \includegraphics[width=0.65\textwidth]{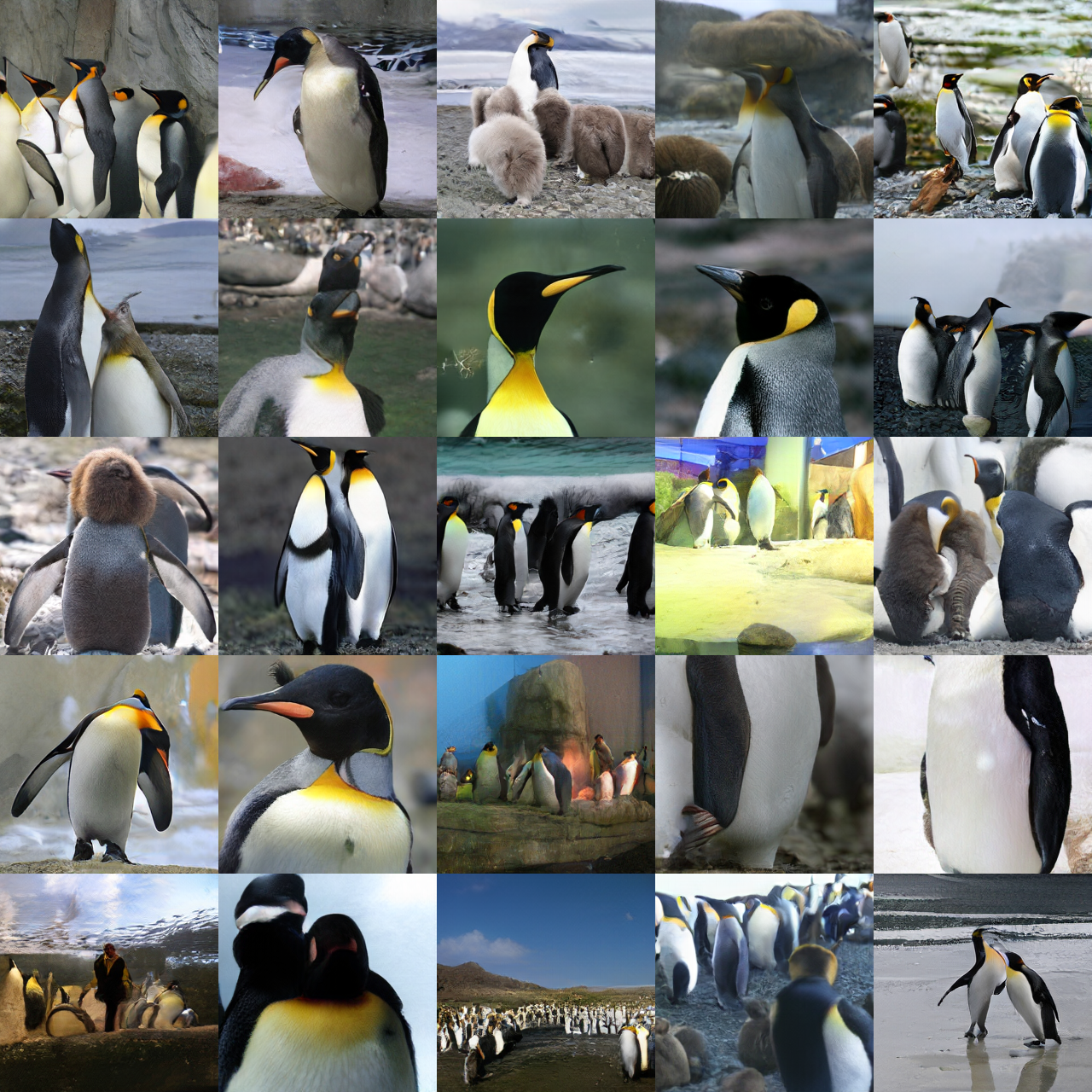}
    \label{fig: gen_c145}
    \caption{king penguin, Aptenodytes patagonica}
\end{figure}

\begin{figure}[H]
    \centering
    \includegraphics[width=0.65\textwidth]{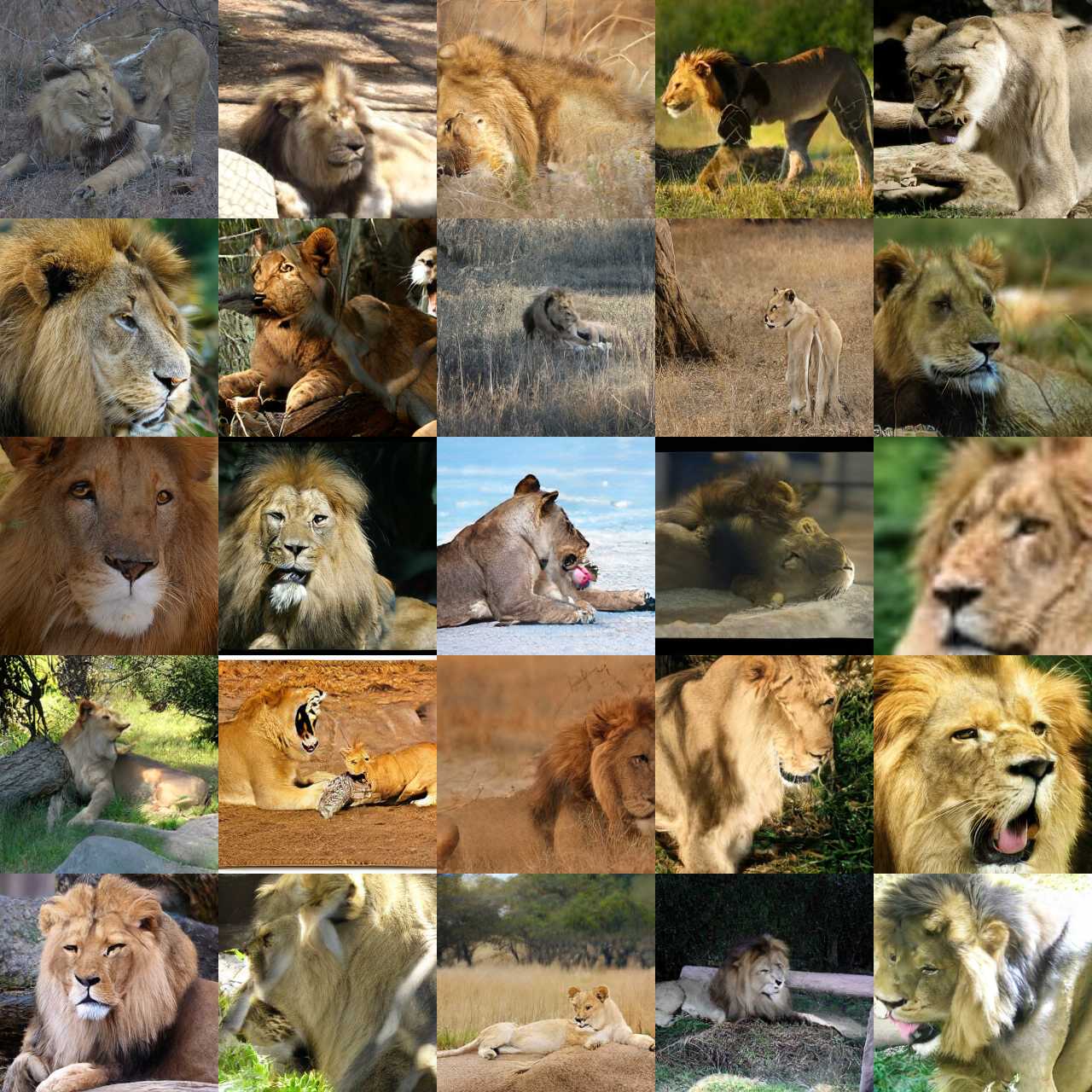}
    \label{fig: gen_c291}
    \caption{lion, king of beasts, Panthera leo}
\end{figure} 

\begin{figure}[H]
    \centering
    \includegraphics[width=0.65\textwidth]{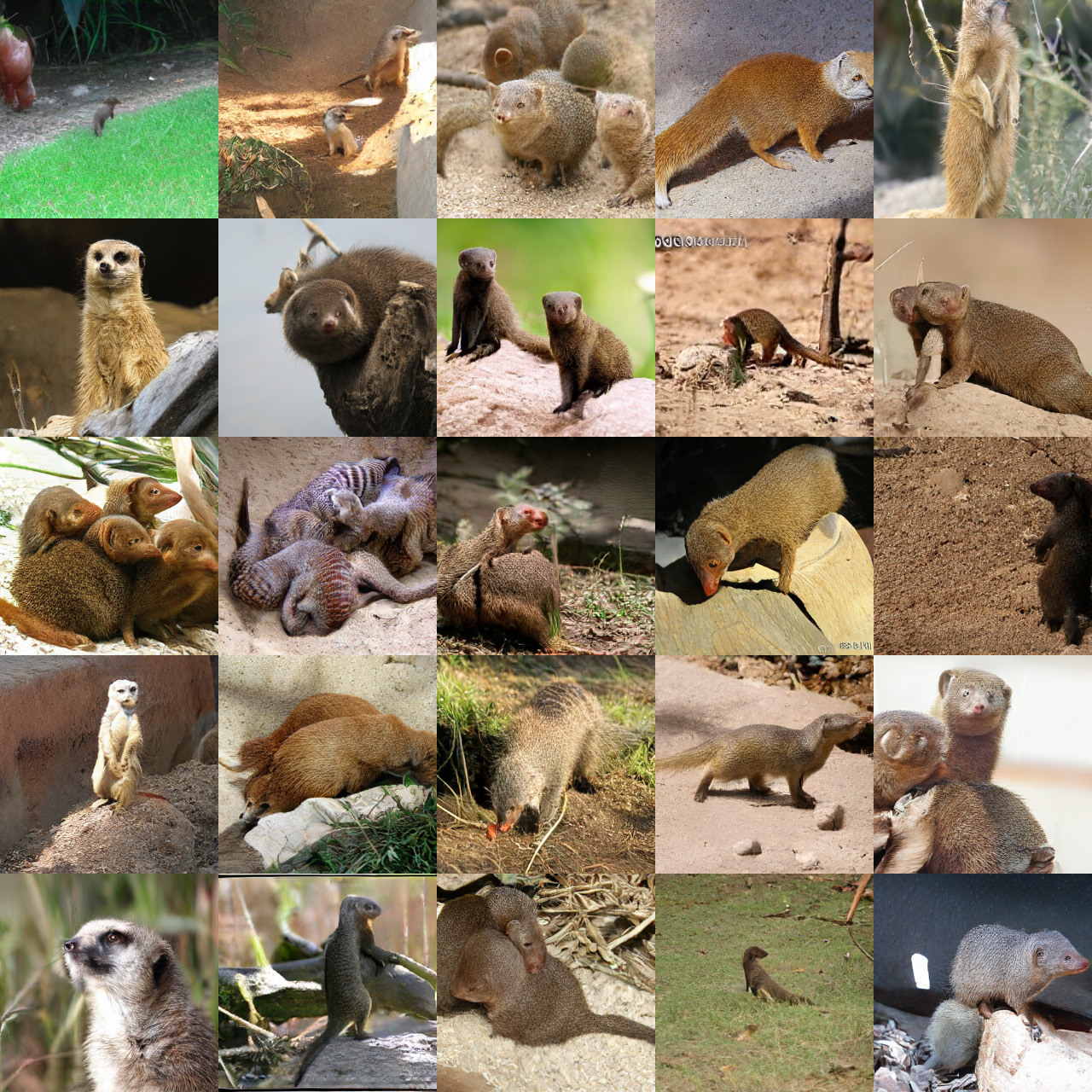}
    \label{fig: gen_c298}
    \caption{mongoose}
\end{figure} 

\begin{figure}[H]
    \centering
    \includegraphics[width=0.65\textwidth]{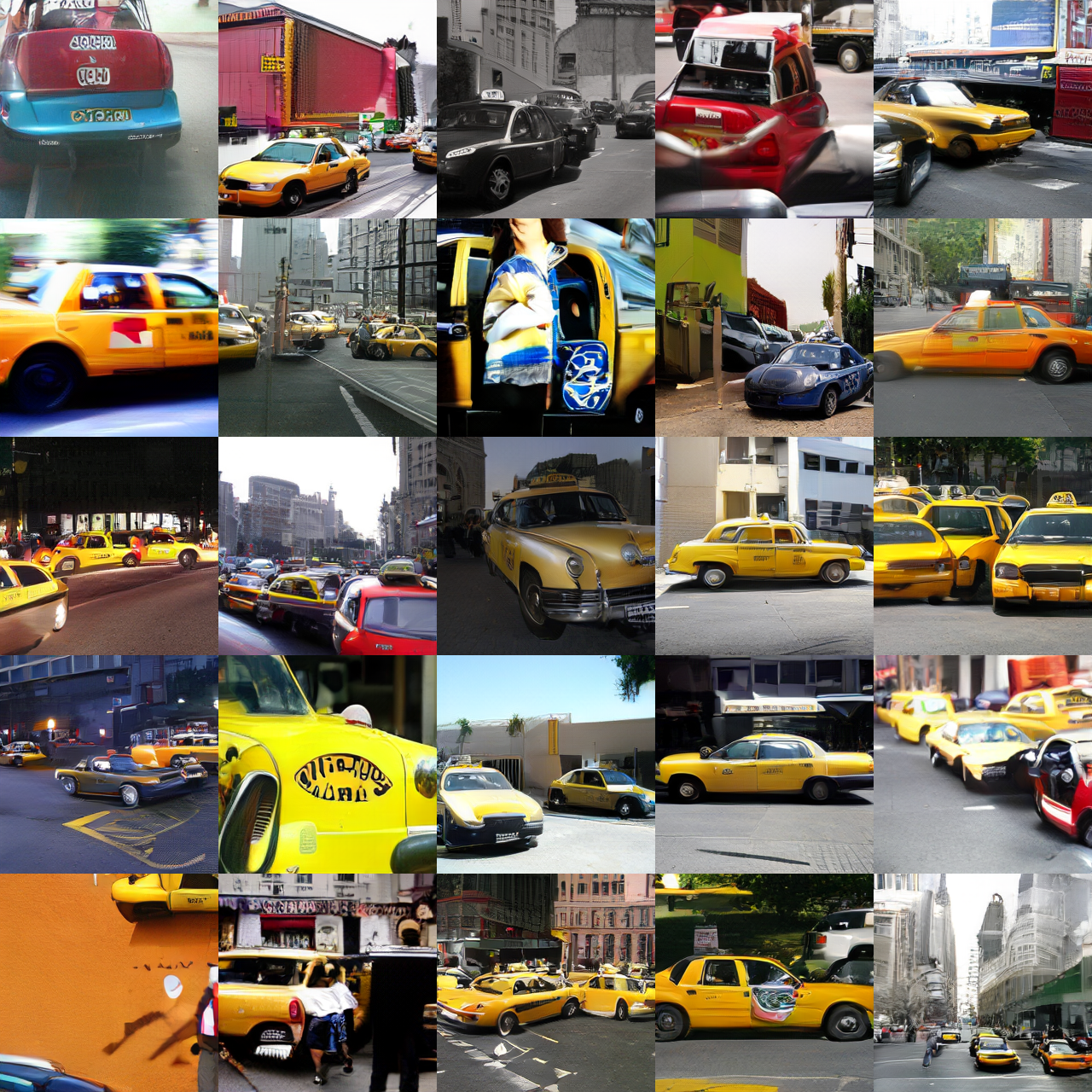}
    \label{fig: gen_c468}
    \caption{cab, hack, taxi, taxicab}
\end{figure} 

\begin{figure}[H]
    \centering
    \includegraphics[width=0.65\textwidth]{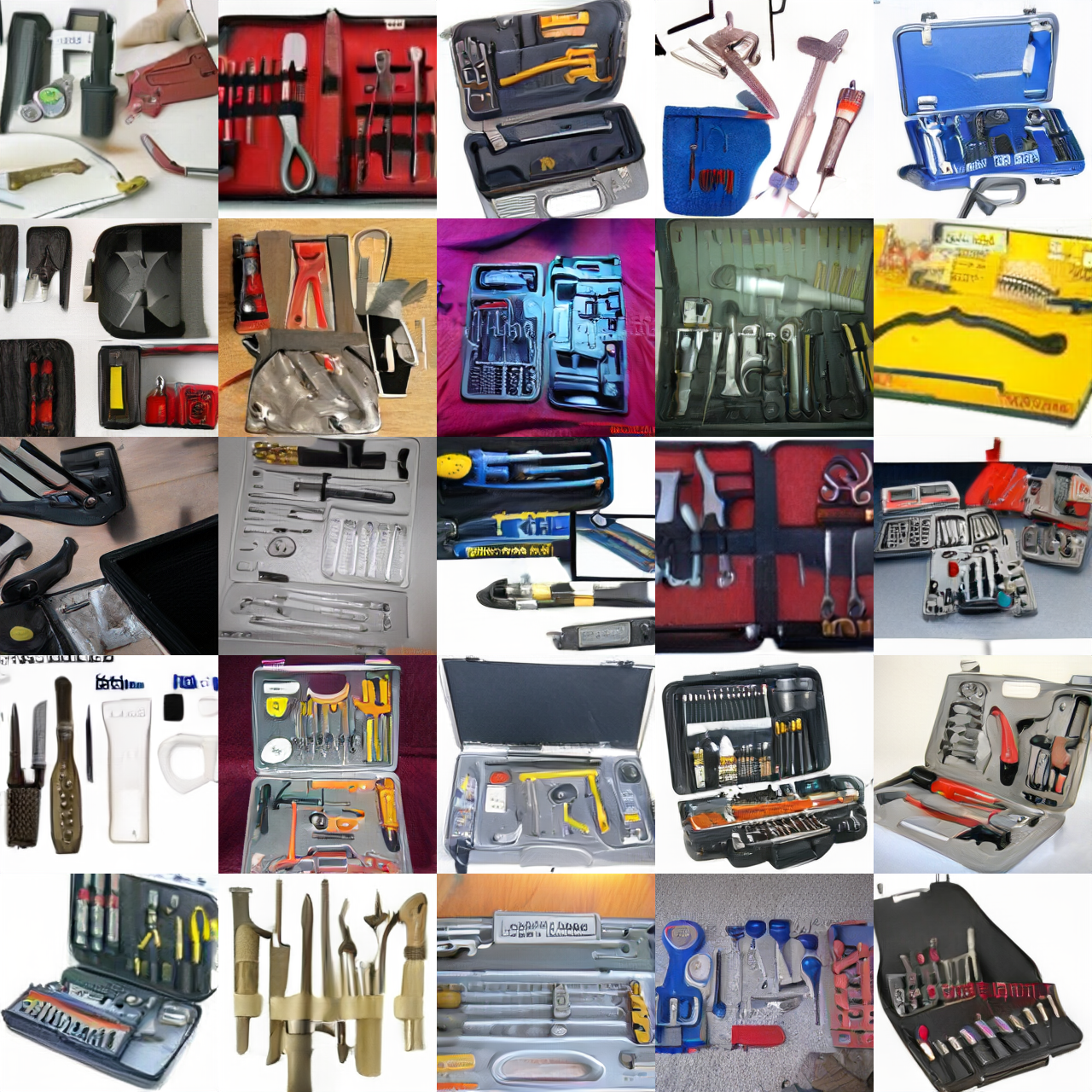}
    \label{fig: gen_c77}
    \caption{carpenter's kit, tool kit}
\end{figure}

\clearpage
\section{Material Tokenization}
\label{app_mattok}

We follow~\cite{liu2026composable} for the data processing, model design and training for crystalline material tokenization using atom only features without using atom neighbor information. Each material is represented by its minimum repeating unit, containing the lattice information as 3 lattice base vectors $L =[\bm{l_1}, \bm{l_2}, \bm{l_3}]$, and a set of atoms within, represented using a one-hot atom type $A$ and continuous 3D coordinate $X=[x, y,z]$ in the fractional coordinate. Following Mattergen~\cite{zeni2025generative}, a normalized rotational invariant representation $\bm{\tilde{L}}$ is used, where $\bm{W}$ and $\bm{V}$ are the left and right singular vectors of $\bm{L}$, respectively, and $\bm{\Sigma}$ is the diagonal matrix of singular values. 

\begin{align*}
    \bm{L} & =\bm{U}\bm{\tilde{L}},\quad\bm{U}=\bm{W}\bm{V}^\top,\quad\bm{\tilde{L}}=\bm{V}\bm{\Sigma}\bm{V}^\top, \\
    \bm{\tilde{L}} & = \frac{\bm{\tilde{L}}}{\sqrt[3]{N_{atom}}}. 
\end{align*}

We used a transformer-based encoder-decoder architecture, where the encoder contains 6 and decoder contains 2 transformer blocks. The hidden dimension is 512 and each transformer block uses 8 attention heads. We set the starting epoch W to 500, and update the codebook at every M = 5 epochs. The model is trained using Adam optimizer, with constant learning rate $1e-4$ and batch size 128. The codebook size K = 8192 and vector size d = 8. The loss function is defined as follows where $\mathcal{L}_{\text{codebook}}$ is enabled only when online correction is used, and $\alpha = 0.01$ is chosen for both VQ-VAE and FlowVQ to keep all loss terms on comparable scales:
\[\mathcal{L}(\theta,\gamma) = \mathcal{L}_{\text{recon}} + \mathcal{L}_{\text{FM}} + \alpha(\mathcal{L}_{\text{codebook}} + \beta \mathcal{L}_{\text{commitment}})\] 
For baseline models, we performed k-means clustering on the training data latent vectors of the autoencoder at epoch W using \texttt{FAISS}~\cite{douze2025faiss}\footnote{\url{https://github.com/facebookresearch/faiss}}, and used the resulting cluster centroids to initialize VQ-VAE and Opt-VQ codebook vectors. To ensure consistency with the initial latent space, and fair comparison from the same starting point, we also initialized the encoder and decoder weights using those of the autoencoder. For LFQ, we used the implementation from \texttt{vector-quantize-pytorch} library\footnote{\url{https://github.com/lucidrains/vector-quantize-pytorch}}.

\section{Action Tokenization}
\label{app_acttok}

The action tokenization is evaluated on the DROID dataset \cite{khazatsky2024droid}, which contains 76k successful demonstration trajectories. We split the original trajectories into 80\% train, 10\% validation and 10\% test set, then further processed them into non-overlapping chunks of 50 steps. The codebook size K is set to 2048 for fair comparison with existing tokenizers, and vector size d = 8. The codebook construction starts at step $W=1000^{th}$, and is update at every $250^{th}$ steps using batch size 512. The model contains 8 transformer blocks for both encoder and decoder, using 8 attention heads and hidden dimension as 512. Similarly, we train the model using a constant learning rate $1e^{-4}$ and Adam optimizer.

For the baseline FAST+ tokenizer, we use the weight downloaded from its original HuggingFace repository\footnote{\url{https://huggingface.co/physical-intelligence/fast}}.


\ifchecklist
    \newpage
    \section*{NeurIPS Paper Checklist}

\begin{enumerate}

\item {\bf Claims}
    \item[] Question: Do the main claims made in the abstract and introduction accurately reflect the paper's contributions and scope?
    \item[] Answer: \answerYes{} 
    \item[] Justification: Our results are backed by experimental evidence provided in section 4 with 4 tasks across 3 domains.

\item {\bf Limitations}
    \item[] Question: Does the paper discuss the limitations of the work performed by the authors?
    \item[] Answer: \answerYes{} 
    \item[] Justification: The limitations are discussed in section 5.

\item {\bf Theory assumptions and proofs}
    \item[] Question: For each theoretical result, does the paper provide the full set of assumptions and a complete (and correct) proof?
    \item[] Answer: \answerNA{} 
    \item[] Justification: This paper does not contain theoretical results.

    \item {\bf Experimental result reproducibility}
    \item[] Question: Does the paper fully disclose all the information needed to reproduce the main experimental results of the paper to the extent that it affects the main claims and/or conclusions of the paper (regardless of whether the code and data are provided or not)?
    \item[] Answer: \answerYes{} 
    \item[] Justification: The experiment details are documented in section 4 and additional hyper-parameter setting in the appendix.

\item {\bf Open access to data and code}
    \item[] Question: Does the paper provide open access to the data and code, with sufficient instructions to faithfully reproduce the main experimental results, as described in supplemental material?
    \item[] Answer: \answerYes{} 
    \item[] Justification: Code is available at: \href{https://github.com/friolero/FlowVQ}{https://github.com/friolero/FlowVQ}.

\item {\bf Experimental setting/details}
    \item[] Question: Does the paper specify all the training and test details (e.g., data splits, hyperparameters, how they were chosen, type of optimizer) necessary to understand the results?
    \item[] Answer: \answerYes 
    \item[] Justification: We provided the details in section 4 under each experiment section.

\item {\bf Experiment statistical significance}
    \item[] Question: Does the paper report error bars suitably and correctly defined or other appropriate information about the statistical significance of the experiments?
    \item[] Answer: \answerYes{} 
    \item[] Justification: We use 50000 images for FID and IS score measure, and report averaged time taken from 5 measures. 

\item {\bf Experiments compute resources}
    \item[] Question: For each experiment, does the paper provide sufficient information on the computer resources (type of compute workers, memory, time of execution) needed to reproduce the experiments?
    \item[] Answer: \answerYes{} 
    \item[] Justification: we included the hardware specification in the experiment section.
    
\item {\bf Code of ethics}
    \item[] Question: Does the research conducted in the paper conform, in every respect, with the NeurIPS Code of Ethics \url{https://neurips.cc/public/EthicsGuidelines}?
    \item[] Answer: \answerYes{} 
    \item[] Justification: We have read the NeurIPS Code of Ethics and made sure that the paper
conforms to it.

\item {\bf Broader impacts}
    \item[] Question: Does the paper discuss both potential positive societal impacts and negative societal impacts of the work performed?
    \item[] Answer: \answerNA{} 
    \item[] Justification: This work is not tied to particular applications but addresses an issue with classic neural tokenization method. The potential societal impacts of certain applications are not relevant to this work.
    
\item {\bf Safeguards}
    \item[] Question: Does the paper describe safeguards that have been put in place for responsible release of data or models that have a high risk for misuse (e.g., pre-trained language models, image generators, or scraped datasets)?
    \item[] Answer: \answerNA{} 
    \item[] Justification: We did not use pretrained models or scraped dataset but only common benchmark datasets.

\item {\bf Licenses for existing assets}
    \item[] Question: Are the creators or original owners of assets (e.g., code, data, models), used in the paper, properly credited and are the license and terms of use explicitly mentioned and properly respected?
    \item[] Answer: \answerYes{} 
    \item[] Justification: Citations to relevant models and works are given.

\item {\bf New assets}
    \item[] Question: Are new assets introduced in the paper well documented and is the documentation provided alongside the assets?
    \item[] Answer: \answerYes{} 
    \item[] Justification: This paper does not introduce new datasets. The details of the models are provided in section 4 and appendix.

\item {\bf Crowdsourcing and research with human subjects}
    \item[] Question: For crowdsourcing experiments and research with human subjects, does the paper include the full text of instructions given to participants and screenshots, if applicable, as well as details about compensation (if any)? 
    \item[] Answer: \answerNA{} 
    \item[] Justification: This work does not involve crowdsourcing experiments.

\item {\bf Institutional review board (IRB) approvals or equivalent for research with human subjects}
    \item[] Question: Does the paper describe potential risks incurred by study participants, whether such risks were disclosed to the subjects, and whether Institutional Review Board (IRB) approvals (or an equivalent approval/review based on the requirements of your country or institution) were obtained?
    \item[] Answer: \answerNA{} 
    \item[] Justification:  This work does not involve human participants.

\item {\bf Declaration of LLM usage}
    \item[] Question: Does the paper describe the usage of LLMs if it is an important, original, or non-standard component of the core methods in this research? Note that if the LLM is used only for writing, editing, or formatting purposes and does \emph{not} impact the core methodology, scientific rigor, or originality of the research, declaration is not required.
    \item[] Answer: \answerNA{} 
    \item[] Justification: The core methodology of this work does not involve LLMs as an important, original, or non-standard component.

\end{enumerate}
\fi

\end{document}